\documentclass[onefignum,onetabnum]{siamonline190516}

\usepackage{graphicx}
\usepackage{subcaption}
\usepackage{tikz} 
\usepackage{amssymb, amsmath, amsfonts}
\usepackage{mathtools}
\usepackage{bbm}
\usepackage{algorithm}
\usepackage[noend]{algpseudocode}
\usepackage{algorithmicx}
\usepackage{color}
\usepackage{multirow}
\usepackage{array}
\usepackage{blkarray}
\usepackage{tabularx}
\usepackage[percent]{overpic} 
\usepackage[usestackEOL]{stackengine}
\usepackage{contour} 
\usepackage{xcolor} 
\usepackage{xfrac} 

\allowdisplaybreaks

\newcommand{\bfh}{\mathbf{h}}
\newcommand{\bfp}{\mathbf{p}}
\newcommand{\bft}{\mathbf{t}}
\newcommand{\bfu}{u}
\newcommand{\bfv}{v}
\newcommand{\bfx}{\mathbf{x}}
\newcommand{\bfy}{\mathbf{y}}
\newcommand{\bfz}{\mathbf{z}}

\newcommand{\norm}[2]{ \left\| #1 \right\|_{#2} }
\newcommand{\indicator}[1]{\mathbbm{1}_{#1}}
\newcommand{\supp}{\operatorname{supp}}

\DeclareMathOperator*{\argmin}{arg\,min}

\def\inpdom{\mathcal{O}}
\def\patch{\mathcal{P}}

\graphicspath{ {./images/} }
\ifpdf
  \DeclareGraphicsExtensions{.eps,.pdf,.png,.jpg}
\else
  \DeclareGraphicsExtensions{.eps}
\fi

\usepackage{enumitem}
\setlist[enumerate]{leftmargin=.5in}
\setlist[itemize]{leftmargin=.5in}

\newsiamremark{remark}{Remark}
\newsiamremark{hypothesis}{Hypothesis}
\crefname{hypothesis}{Hypothesis}{Hypotheses}
\newsiamthm{claim}{Claim}

\headers{A nonlocal feature-driven exemplar-based approach for image inpainting}{V. Reshniak, J. Trageser, and C.G. Webster}

\title{A nonlocal feature-driven exemplar-based approach for image inpainting\thanks{Submitted to the editors DATE
\funding{\newline This manuscript has been co-authored by UT-Battelle, LLC, under contract DE-AC05-00OR22725 with the US Department of Energy (DOE). The US government retains and the publisher, by accepting the article for publication, acknowledges that the US government retains a nonexclusive, paid-up, irrevocable, worldwide license to publish or reproduce the published form of this manuscript, or allow others to do so, for US government purposes. DOE will provide public access to these results of federally sponsored research in accordance with the DOE Public Access Plan (http://energy.gov/downloads/doe-public-access-plan) \newline Sandia National Laboratories is a multimission laboratory managed and operated by National Technology \& Engineering Solutions of Sandia, LLC, a wholly owned subsidiary of Honeywell International Inc., for the U.S. Department of Energy’s National Nuclear Security Administration under contract DE-NA0003525. This paper describes objective technical results and analysis. Any subjective views or opinions that might be expressed in the paper do not necessarily represent the views of the U.S. Department of Energy or the United States Government.}}}

\author{ 
     Viktor Reshniak\thanks{Computational and Applied Mathematics, Oak Ridge National Laboratory, Oak Ridge, TN, 37831-2008 (\email{reshniakv@ornl.gov}).}
\and Jeremy Trageser\thanks{Center for Computing Research, Sandia National Laboratories, Albuquerque, NM, 87185-1322 (\email{jtrages@sandia.gov})}
\and Clayton G.~Webster\thanks{Department of Mathematics, University of Tennessee at Knoxville (\email{cwebst13@utk.edu}) and Behavioral Reinforcement Learning Lab (BReLL), Lirio LLC, Knoxville, TN 37923 (\email{cwebster@lirio.co}).}
  }

\usepackage{amsopn}

\DeclareMathSymbol{\shortminus}{\mathbin}{AMSa}{"39}

\ifpdf
\hypersetup{
  pdftitle={A nonlocal feature-driven exemplar-based approach for image inpainting},
  pdfauthor={V. Reshniak, J. Trageser, and C.G. Webster}
}
\fi

\begin{document}

\maketitle

\begin{abstract}
    We present a nonlocal variational image completion technique which admits simultaneous inpainting of multiple structures and textures in a unified framework.
    The recovery of geometric structures is achieved by using general convolution operators as a measure of behavior within an image.
    These are combined with a nonlocal exemplar-based approach to exploit the self-similarity of an image in the selected feature domains and to ensure the inpainting of textures.
    We also introduce an anisotropic patch distance metric to allow for better control of the feature selection within an image and present a nonlocal energy functional based on this metric.
    Finally, we derive an optimization algorithm for the proposed variational model and examine its validity experimentally with various test images.
\end{abstract}



\section{Introduction}
\label{intro}

Inpainting is a process through which lost, deteriorated, or undesirable portions of images and videos may be replaced with a visually plausible interpolation. The applications are plentiful and inpainting is an active area of research in image processing. The majority of image inpainting methods can be categorized into two groups: geometry-oriented and texture-oriented.

In geometry-oriented inpainting, such as in \cite{ballester2001filling,masnou1998level,shen2002mathematical,shen2003euler}, some regularity in a functional representation of the image is assumed. The inpainting takes place by interpolating from the boundary of the inpainting region with the regularity assumptions imposed on the image. 
This is typically accomplished by solving a boundary-value partial differential equation, where the the solution domain is the inpainting region and boundary is the set of pixels immediately surrounding it. 
These methods have been quite successful when the inpainting region is small, such as when removing scratches, text, or minor blemishes \cite{Bertalmio:2000,shen2002mathematical}. 
Geometric objects such as curves have also been successfully propagated into the inpainting region in numerous works, e.g. \cite{shen2003euler,Cao2011}. Due to the fact that typically only the pixels immediately surrounding the inpainting region are utilized to inform the inpainting algorithm in geometry-oriented methods, potentially useful information not in proximity to the inpainting domain is frequently neglected.

Alternatively, in many texture-oriented methods, the goal is to exploit the self-similarity inherent in many images. The value at a pixel in the inpainting region is typically determined by comparing the neighborhood around the pixel, often called a patch, and the neighborhoods of pixels in the intact/desirable region of the image. When the patches are determined to be similar with respect to some criterion, the pixel in the inpainting region is assigned characteristics from pixels in the intact/desirable region of the image. These approaches are often referred to as exemplar-based methods and nonlocality is an intrinsic property. Early examples of such methods are explored in \cite{Bornard2002,Criminisi2003,Drori2003}. Subsequent research and development of exemplar-based methods has been extensive with many important contributions such as  \cite{Gilboa2008,Arias2011}. Exemplar-based methods are celebrated for their ability to recover textures and repetitive structures; however, recreating geometry when a specific example of the structure is not in the known image is limited and not well-understood \cite{Arias2011}.

A synthesis of these two paradigms has been considered in multiple works such as \cite{Cao2011,Arias2011}; however, many such algorithms are driven by similarities in pixel magnitudes rather than underlying features. Many images have self-similarity in features that would be ignored by an inpainting technique that focused solely on differences in pixel color. For example, in an image with a black cat and a white cat, one could utilize the self-similarity in structure and texture in one cat to reconstruct a region missing from the other cat. In the continuum (non-discrete) framework, frequently derivatives are the tools employed to measure and understand function behavior. When discontinuous or discrete systems are considered such as in images and data sets, classical derivatives are not well-defined; however, generalizations of such operations exist in the form of nonlocal operators \cite{du2013nonlocal,DINEZZA2012521}. Much like their local counterparts, the nonlocal operators can provide measures of structure and texture on discontinuous or discrete domains. In this work we utilize general nonlocal operators to explore a method of exemplar-based inpainting executed in selected feature domains.

Introducing nonlocal operators into the image processing field has been immensely successful, e.g. nonlocal total variation \cite{Gilboa2008} and nonlocal means \cite{Buades2005}; however, even in these seminal works, similarity comparisons between patches are largely dependent on color or intensity differences rather than true textural and structural differences. 
As a potential remedy, in \cite{Arias2011} a framework capable of considering structural and textural differences was developed.
Their method is a generalization of the method proposed in \cite{Gilboa2008}, although their method was a marked improvement as it did not require a priori determining which pixels were self-similar. 
In addition, rather than solely considering patch differences based entirely on pixel color/intensity, patch differences dependent on the gradient of the image were additionally utilized. The use of gradients in the inpainting process was later explored in \cite{Liu2012,Newson2014}. In this work we delve further into the framework introduced in \cite{Arias2011} and consider a general convolution operator which allows us to focus on the types of distinguishing features, such as edges or textures, we wish to utilize in the inpainting process. Employing a more general feature space in the image reconstruction allows the inpainting algorithm to take advantage of additional self-similarity within the image that may be ignored by more traditional filters.

The organization of this paper is as follows. 
In Section \ref{ref:problem} we formulate the image inpainting problem and review various inspirational works for the model developed in this paper. 
In Section \ref{sec:model} we present our model as well as an algorithm with which it may be solved. 
We continue in Section \ref{sec:examples} with examples of our model applied to various damaged images. 
We conclude our discussion with Section \ref{sec:conclusion}.

\section{Problem formulation and related works}\label{ref:problem}

In this section we formalize the image inpainting problem and review several inspirational works for the model developed in this paper.

\begin{figure}
    \stackinset{c}{0pt}{b}{-15pt}{\small Original image}{\includegraphics[width=0.24\textwidth]{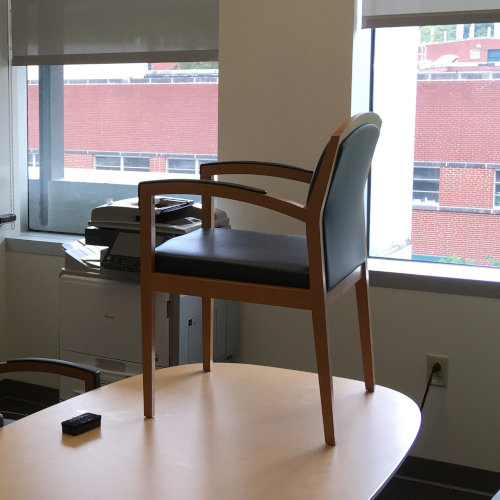}}
    \stackinset{c}{0pt}{b}{-15pt}{\small Inpainting region}{\includegraphics[width=0.24\textwidth]{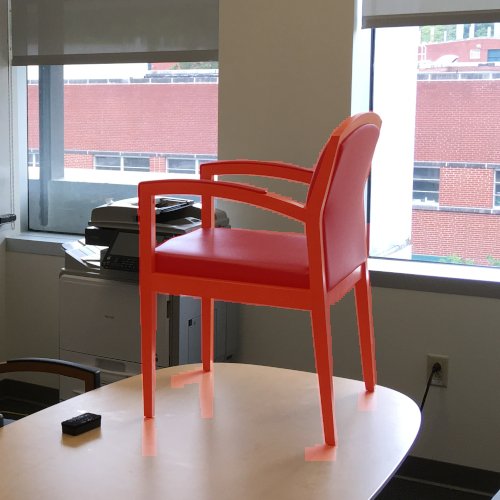}}
    \stackinset{c}{0pt}{b}{-30pt}{\small \Centerstack{Image domain $\Omega$ with \\ inpainting region $\mathcal{O}$}}{
    \begin{overpic}[width=0.24\textwidth,tics=5]{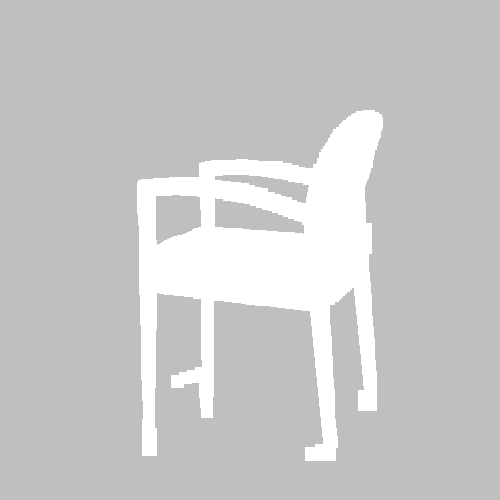}
        \put (50,45) {\scriptsize $\inpdom$}
        \put (10,85) {\scriptsize $\inpdom^c:=\Omega\setminus\inpdom$}
    \end{overpic}}
    \stackinset{c}{0pt}{b}{-15pt}{\small Inpainted image}{\includegraphics[width=0.24\textwidth]{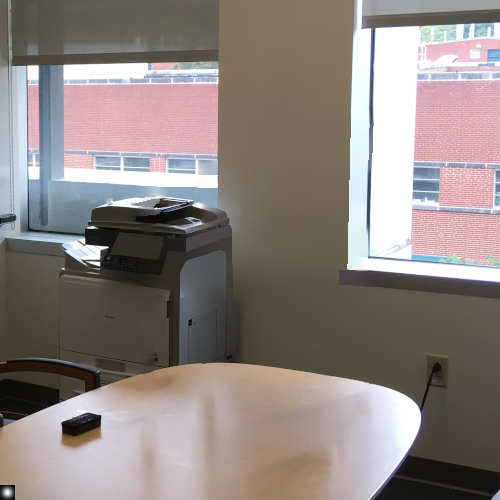}}
    \caption{Example of an image inpainting problem.}
    \label{fig:inp_domain}
\end{figure}

We begin by defining $\Omega \subset \mathbb{R}^2$ to be an image domain on which we define a function $\hat{\bfu}: \Omega \rightarrow [0,1]$, corresponding to the initial image\footnote{Color images should treat functions mapped into $[0,1]^3$; however, the majority of inpainting methods are defined for scalar-valued functions and merely treat each color channel separately and then compile the result.}. 
In the inpainting problem, one typically desires to provide a plausible replacement of $\hat{\bfu}$ on an unknown or undesirable region, $\mathcal{O} \subset \Omega$, of an image. 
We refer to $\mathcal{O}$ as the inpainting region of the image defined on $\Omega$. Filling in the inpainting region is most certainly an ill-posed problem as there are typically multiple plausible ways to fill in the missing part of the image. 
For example, in Figure~\ref{fig:inp_domain}, a region containing the entire chair is removed from the image and one could inpaint the unknown/removed part with a chair with any number of color possibilities or even with no chair at all. Consequently, the goal of inpainting is not a true recovery of the inpainting region, but rather to produce a satisfactory replacement. 
More precisely, the inpainting problem is to elicit a function $\bfu:\Omega \rightarrow [0,1]$ such that $\bfu \approx \hat{\bfu}$ for $\bfx \in \mathcal{O}^c = \Omega \backslash \mathcal{O}$, and $\bfu$ assumes values within the inpainting region $\mathcal{O}$ which attains a ``reasonable" replacement of $\hat{\bfu}$.

While there are a multitude of methods for inpainting, variational frameworks are particularly popular, e.g.,  \cite{Arias2011,ballester2001filling,Gilboa2007,masnou1998level,mumford1989optimal,Osher2017,shen2002mathematical}. In many variational frameworks, the solution $\bfu$ to the inpainting problem is obtained through the minimization of some functional $\mathcal{E}[\bfu]$ over an appropriate admissibility set $U$. Various models are distinguished by the form of the functional $\mathcal{E}[\bfu]$ and the admissibility set $U$. In order to lend credibility to the model proposed in this work as well as acknowledge several inspirational works, we now review several well-known variational models.

\textbf{Local Approaches}:
In local variational approaches, some regularity in the functional representation of the image is assumed. In these approaches, global features and pattern recognition are not utilized. These methods tend to perform well when structure needs to be diffused into the region; however, they fail to take into account symmetry, such as mirror symmetry in a face, and repetitive features, such as patterns in fabric, within an image. 
It is well known that these local approaches are typically more successful when the inpainting region is small or the region to be recovered is sufficiently regular, but fail to recover texture \cite{Cao2011}.  

\underline{Local total variation}: In \cite{shen2002mathematical} a PDE inpainting procedure, often referred to as the TV inpainting algorithm, was developed based on the popular total variation (TV) denoising algorithm of Rudin, Osher, and Fatemi (the ROF Model) \cite{Rudin1992}. In the total variation inpainting model, they introduce a region $E$ about the inpainting region $\mathcal{O}$ such that $E \subset \mathcal{O}^c$ and $\partial \mathcal{O}$ is contained in the interior of $\mathcal{O} \cup E$. The functional proposed in \cite{shen2002mathematical} is
\begin{align}\label{eq:ROF}
    \mathcal{E}[\bfu] :=  \| \nabla \bfu \|_{L^1(\mathcal{O} \cup E)} +  \frac{\lambda}{2}  \| \hat{\bfu}- \bfu \|_{L^2(E)}^2,
\end{align}
where $\lambda>0$ is a scaling constant and $\nabla$ represents the gradient operator. The functional $\mathcal{E}[\bfu]$ is minimized over the admissibility set $U = BV(\mathcal{O} \cup E)$, i.e., the set of functions of bounded variation on $\mathcal{O}\cup E$. The $\| \hat{\bfu}- \bfu \|_{L^2(E)}^2$ term is sometimes referred to as a fidelity term and prevents the recovered image from diverging too significantly from the initial image in the desirable/intact region $E$. Under the minimization of \eqref{eq:ROF}, the $\| \nabla \bfu \|_{L^1(\mathcal{O}\cup E)}$ term will reduce the total variation over the region $\mathcal{O} \cup E$ and consequently diffuses structure into the inpainting region $\mathcal{O}$ from the region $E$. The local total variation inpainting model is computationally inexpensive, but suffers from various shortcomings such as the production of artifacts like staircasing. In an attempt to mitigate or eliminate these artifacs, several higher-order models were developed, e.g. \cite{bertozzi2007analysis,burger2009cahn,papafitsoros2014combined,shen2003euler}.  

\underline{Local first and second-order total variation}: In \cite{papafitsoros2014combined} a higher-order generalization of the local total variation model \eqref{eq:ROF} was proposed. As in the total variation model, a collar domain $E$ about the inpainting region $\mathcal{O}$ is utilized. The proposed functional in \cite{papafitsoros2014combined} is
\begin{align}\label{eq:ROFGeneralized1}
    \mathcal{E}[\bfu] :=
    \alpha \int\displaylimits_{\mathcal{O} \cup E} f(\nabla \bfu) d \bfx + \beta \int\displaylimits_{\mathcal{O} \cup E} g(\nabla^2 \bfu) d \bfx 
    +  \frac{\lambda}{2}  \| \hat{\bfu}- \bfu \|_{L^2(E)}^2,
\end{align}
where $\alpha$ and $\beta$ are non-negative regularization parameters, $\nabla^2$ is the Hessian operator, and $f$ and $g$ are convex nonnegative functions with at most linear growth at infinity. The functional \eqref{eq:ROFGeneralized1} is minimized over the space of bounded Hessian $BH(\Omega)$. If one takes $\beta = 0$ and $f = | \cdot |$, then \eqref{eq:ROFGeneralized1} reduces to \eqref{eq:ROF}. As was demonstrated in \cite{papafitsoros2013combined,papafitsoros2014combined}, introducing higher-order terms alleviates the staircasing effect produced by the local total variation model. The associated cost of these improvements is some blurring, which can be controlled to some extent by the parameter $\beta$, as well as increased computational cost.

As mentioned previously, local variational approaches tend to be most successful when the inpainting region is thin or small. When the impainting region is larger or texture recovery is important, so-called nonlocal approaches often offer superior performance.

\textbf{Nonlocal Approaches}:
In nonlocal variational approaches, rather than just propagating structure into the inpainting region through regularity restrictions, global features and pattern recognition are often exploited. In order to take advantage of repetitive features such as texture about a point, the neighborhood of a point, often referred to as a patch, is utilized. To formalize the concept of patches, let $\patch$ be the rectangle centered at $(0,0)$ with side lengths $s_1$ and $s_2$, i.e., 
\begin{equation}\label{eqn:patchdef}
\patch := \left\{(x,y) \in \mathbb{R}^2: |x| < \frac{s_1}{2}, |y| < \frac{s_2}{2} \right\}.
\end{equation}
We define the patch of $\bfu$ about the point $\bfx$ as $ p_\bfu(\bfx) := \bfu \left( \bfx + \patch \right),$
where $\bfx + \patch = \left\{\bfx + \bfy : \bfy \in \patch \right\}$ denotes the Minkowski summation.
Nonlocal models view an image $\bfu$ as a manifold $\mathcal{M}_\bfu$ embedded in the space of all patches of $\bfu$, $\left\{ p_\bfu(\bfx) : \bfx \in \Omega \right\}$. Nonlocal methods frequently take advantage of the low dimensionality inherent in the patch manifold \cite{Osher2017}. Discrete representations of patch manifolds are often interpreted as weighted graphs and the choice of a weighting function determines the flow of information from the known to the unknown regions of the image. 

\underline{Nonlocal total variation}: A nonlocal variant of the local total variation inpainting model was introduced in \cite{Gilboa2008}. In this model, a nonlocal gradient is employed as a regularization term instead of the local derivatives in \eqref{eq:ROF}.
In the nonlocal total variation framework, both isotropic and anisotropic functionals are considered:
\begin{subequations}\label{eqn:NLTVFunctionals}
    \begin{align}
        \label{eqn:isoNLTV}
        \mathcal{E}[\bfu] &:= \int\limits_{\Omega} \sqrt{ \int\limits_{\Omega} w(\bfx,\bfy) \big( \bfu(\bfy)-\bfu(\bfx) \big)^2 d \bfy } d \bfx 
        + \lambda \int\limits_{\mathcal{O}^c} \big( \hat{\bfu} - \bfu \big)^2 d \bfx; \text{ and}
        \\
        \label{eqn:anisoNLTV}
        \mathcal{E}[\bfu] &:= \int\limits_{\Omega} \int\limits_{\Omega} \sqrt{w(\bfx,\bfy)} \big| \bfu(\bfy)-\bfu(\bfx) \big| d\bfy d\bfx 
        + \lambda \int\limits_{\mathcal{O}^c} \big( \hat{\bfu} -  \bfu \big)^2 d \bfx.
    \end{align}
\end{subequations}
respectively.
The weighting function $w(\bfx,\bfy)$ in \eqref{eqn:NLTVFunctionals} is given by
\begin{align}\label{eqn:GilboaWeights}
    w(\bfx,\bfy) = 
    \begin{cases}
        1, &\bfy \in A_n(\bfx) \text{ or } \bfx \in A_n(\bfy),
        \\
        0, &\text{otherwise},
    \end{cases}
\end{align}
where for a given $n$, $A_n(\bfx)$ is the set of $n$ nearest neighbors of $\bfx$ in a weighted patch distance metric $d[\hat{\bfu}](\bfx,\cdot)$ based on the initial image $\hat{\bfu}$:
\begin{align}\label{eq:patch_dist_1}
    d[\hat{\bfu}](\bfx,\bfy) &= \int\limits_{\patch} \big| \hat{\bfu}(\bfy+\bfh) - \hat{\bfu}(\bfx+\bfh) \big|^2 d\patch(\bfh),
\end{align}
with $d\patch(\bfh) \simeq \exp \left( -\norm{\bfh}{2}^2 \right) d\bfh$.
In this formulation, the weighting function $w(\bfx,\bfy)=1$ when the patches around $\bfx$ and $\bfy$ are more similar and $w(\bfx,\bfy) = 0$ when the patches around $\bfx$ and $\bfy$ are less similar. 
Consequently, under the minimization of the functional $\mathcal{E}[\bfu]$ in \eqref{eqn:isoNLTV} or \eqref{eqn:anisoNLTV}, the regularization term (the first term in \eqref{eqn:isoNLTV} or \eqref{eqn:anisoNLTV}) will cause $\bfu(\bfy)$ and $\bfu(\bfx)$ to be more similar when their corresponding patches are similar with respect to \eqref{eq:patch_dist_1}. 
The second term in \eqref{eqn:isoNLTV} or \eqref{eqn:anisoNLTV} is the fidelity term which forces the recovered image $\bfu$ to be close to the initial image $\hat{\bfu}$ in the complement of the inpainting region, $\mathcal{O}^c$. 
One of the most serious critiques of this model is that the weighting function $w(\bfx,\bfy)$ is not adaptive and so the inpainting algorithm requires one to a priori know the weighting function $w$ before minimizing the functional. 
This is a significant drawback when the inpainting region covers a considerable portion of the neighborhood of a pixel. 

\underline{Correspondence map}: Early examples of adaptivity in the weighting function can be found in correspondence map inpainting models. In these models, a correspondence map $\Gamma: \mathcal{O} \rightarrow \Omega \backslash \mathcal{O}$ is determined. Given a pixel $\bfx$ and its image $\Gamma(\bfx)$ in the correspondence map, the image value $\bfu(\bfx)$ is replaced with the image value $\hat{\bfu} \left( \Gamma(\bfx) \right)$.
An early correspondence map inpainting model was proposed in \cite{Demanent2003} where the functional
\begin{align}\label{eqn:Demanetmodel}
    \mathcal{E}[\Gamma] := \int\limits_{\mathcal{O}} \int\limits_{\patch} | \hat{\bfu}(\Gamma(\bfx+\bfh)) - \hat{\bfu}(\Gamma(\bfx)+\bfh) |^2 d\bfh d\bfx
\end{align}
was introduced.
Minimizing \eqref{eqn:Demanetmodel} over $\Gamma$ produces $\bfu(\bfx) = \hat{\bfu}(\Gamma(\bfx))$ for all $\bfx \in \mathcal{O}$. Unfortunately, \eqref{eqn:Demanetmodel} is highly non-convex and cannot be minimized easily \cite{Aujol2010}. In order to overcome this obstacle, in \cite{Wexler2007,Kawai2009} a relaxed variant of \eqref{eqn:Demanetmodel} was produced by adding the unknown image $\bfu$ to the functional:
\begin{align}\label{eq:Wexler_model}
    \mathcal{E}[\bfu,\Gamma] := \int\limits_{\mathcal{\tilde{O}}} \int\limits_{\patch} | \bfu(\bfx+\bfh) - \hat{\bfu}(\Gamma(\bfx)+\bfh) |^2 d\bfh d\bfx,
\end{align}
where $\tilde{\mathcal{O}} := \mathcal{O} + \patch = \{ \bfx \in \Omega : (\bfx + \patch) \cap \mathcal{O} \neq \emptyset \}$ is the set of pixels whose corresponding patches intersect the inpainting domain $\mathcal{O}$. 

\underline{Adaptive weight graph-based regularization}: A generalization of the nonlocal total variation model which employs adaptive weighting was introduced in \cite{Peyre2011}. This model tackled the issue of assigning weights between pixels when one or both of the corresponding patches intersects the inpainting region. In addition to the novelty of an adaptive weighting scheme, the model also regularizes on patches rather than on pixels; this consideration improves convergence properties and stability within their proposed algorithm. The functional proposed in \cite{Peyre2011} is 
\begin{equation}\label{eqn:graphbasedfunctional}
    \mathcal{E}[u] := \int\displaylimits_{\Omega} \int\displaylimits_{\Omega} w_{\bfu} (\bfx,\bfy) d_{\bfu} (\bfx,\bfy) d \bfx d \bfy + \lambda \int\displaylimits_{\mathcal{O}^c} (\hat{\bfu}-\bfu)^2 d \bfx
\end{equation}
with $d_{\bfu}(\bfx,\bfy)$ representing the patch distance metric ({\em cf}. \eqref{eq:patch_dist_1}) 
\begin{equation*}
    d_\bfu (\bfx,\bfy) = \int\limits_{\mathcal{P}} |\bfu(\bfy+\bfh)-\bfu(\bfx+\bfh)|^2 d \bfh
\end{equation*}
and the adaptive weighting function $w_{\bfu}(\bfx,\bfy)$ described~by
\begin{align*}
    w_\bfu(\bfx,\bfy) \simeq \exp \left( -d_\bfu(\bfx,\bfy) \right).
\end{align*}
Since these weights are functions of the optimal solution $\bfu$, the algorithm is adaptive to the image content. By introducing adaptive weights, the nonlocal functional \eqref{eqn:graphbasedfunctional} is nonconvex in $\bfu$ and is more expensive and challenging to minimize than the nonlocal total variation function \eqref{eqn:NLTVFunctionals}. However, the resulting optimization problem can be solved by the alternating minimization technique described in \cite{Peyre2011}.

\underline{Higher-order adaptive graph-based regularization: } 
The final model we review was proposed in \cite{Arias2011}, wherein a variational framework which employs higher-order graph regularization techniques was introduced by means of the functional
\begin{align}\label{eq:Arias_model}
    \mathcal{E}[\bfu, w] &:=
    \frac{1}{\sigma} \int\limits_{\tilde{\mathcal{O}}} \int\limits_{\tilde{\mathcal{O}}^c} w(\bfx,\bfy) d[\bfu, \nabla \bfu](\bfx,\bfy) d\bfy d\bfx
    + \int\limits_{\tilde{\mathcal{O}}} \int\limits_{\tilde{\mathcal{O}}^c} w(\bfx,\bfy) \log w(\bfx,\bfy) d\bfy d\bfx,
    \\ \nonumber
    &\text{subject to } \quad
    \int\limits_{\tilde{\mathcal{O}}^c} w(\bfx,\bfy) d\bfy = 1,
\end{align}
where $\tilde{\mathcal{O}} := \mathcal{O} + \mathcal{P}$ and $d[\bfu,\nabla \bfu](\bfx,\bfy)$ is a higher-order weighted patch distance metric given by
\begin{align}\label{eq:patch_dist_grad}
    d[\bfu,\nabla \bfu](\bfx,\bfy) 
    = \int\limits_{\patch} \Big( \lambda |\bfu(\bfy+\bfh) - \bfu(\bfx+\bfh)|^q
    + (1-\lambda) \norm{\nabla \bfu(\bfy+\bfh) - \nabla \bfu(\bfx+\bfh)}{r}^r \Big) d\patch(\bfh)
\end{align}
with $r,q \in \{1,2\}$ and $\lambda \in [0,1]$. Unlike several of the previous models, \eqref{eq:Arias_model} does not employ a fidelity term. Rather, the model relies on the entropy term 
\begin{equation*}
    - \int\limits_{\tilde{\mathcal{O}}^c} w(\bfx,\bfy) \log w(\bfx,\bfy) d\bfy,
\end{equation*}
which was justified in \cite{Arias2011} by appealing to the principle of maximum entropy \cite{Jaynes1957}. In order to determine the form of the weighting function $w(\bfx,\bfy)$, the function $\bfu$ is held fixed and \eqref{eq:Arias_model} is minimized in $w(\bfx,\bfy)$. The resulting weighting function is a probability distribution over $\mathcal{\tilde{O}}^c$ given by
\begin{align}\label{eq:Arias_weights}
    w(\bfx,\bfy) = \frac{ \exp \left( - d[\bfu,\nabla \bfu](\bfx,\bfy)/\sigma \right)}{{\int\limits_{\tilde{\inpdom}_*^c} \exp \left( -d[\bfu,\nabla \bfu](\bfx,\bfy)/\sigma \right) d\bfy}}.
\end{align} 
Here $\sigma$ is a parameter which controls the quantity of patches considered in the comparison.
With the weighting function described by \eqref{eq:Arias_weights}, the functional \eqref{eq:Arias_model} is a generalization of the adaptive weight graph-based regularization functional \eqref{eqn:graphbasedfunctional}. The parameter $\sigma$ in \eqref{eq:Arias_weights} determines the selectivity of the weights. When $\sigma \to \infty$, the weighting function $w(\bfx,\bfy)$ tends to the uniform distribution over~$\mathcal{\tilde{O}}^c$. Alternatively, when $\sigma \to 0$, the weighting function $w(\bfx,\bfy)$ tends to Dirac deltas and the functional in \eqref{eq:Arias_model} is a direct generalization of \eqref{eq:Wexler_model}. In order to minimize the functional \eqref{eq:Arias_model} with weighting function \eqref{eq:Arias_weights}, an iterative approach of alternating between the updating of $\bfu$ and $w(\bfx,\bfy)$ may be utilized. This process is illustrated in \cite{Arias2011}.

The model in \cite{Arias2011} has two main novelties. The first novelty is the justification of the exponential form of the adaptive weighting function \eqref{eq:Arias_weights} through the entropy term in \eqref{eq:Arias_model}. The second novelty is the inclusion of the gradient in the functional in \eqref{eq:Arias_model}. The addition of the gradient produces smoother continuation of information across the boundary of the inpainting region. In addition, the gradient allows filtration of additive differences in brightness from the patch difference metric \eqref{eq:patch_dist_grad}, permitting less emphasis to be placed on differences in color or intensity when comparing patches.

\section{Our approach}
\label{sec:model}

One of the contributions of this work is the introduction of a model which permits simultaneous utilization of multiple features to guide the inpainting procedure.
The proposed model is a generalization of many of the frameworks described in Section \ref{ref:problem}. We begin with the introduction of a generalized anisotropic patch distance metric based on general convolution operators:
\begin{align}\label{eq:our_dist}
    &d[\bfu,g](\bfx,\bfy) =
    \sum_{i=1}^{N_{g}} \int\displaylimits_{\patch} \lambda_i(\bfx+\bfh) \Big| (g_i*\bfu)(\bfx+\bfh) - (g_i*\bfu)(\bfy+\bfh) \Big|^2 d\patch(\bfh),
\end{align}
where each $g_i$ is a kernel of a scalar-valued convolution operator, which we henceforth refer to as a filter, that determines the feature(s) of interest, $N_g$ is the quantity of filters, and anisotropic coefficients $\lambda_i(\bfx)$ determine the degree of dominance of each filter $g_i$ at a pixel~$\bfx$. 
If the feature(s) described by $g_i$ are similar (dissimilar) between two patches at $\bfx$ and $\bfy$, then the metric described by \eqref{eq:our_dist} is small (large). 
Note that the distance metric \eqref{eq:our_dist} is a direct generalization of \eqref{eq:patch_dist_grad} with $r = q = 2$. 
As opposed to the models discussed in Section \ref{ref:problem}, by employing the more general formulation in \eqref{eq:our_dist}, the similarity comparison can depend on multiple targeted features.
Moreover, the introduction of the anisotropic coefficients $\lambda_i(\bfx)$ allows for better local control of the feature selection.
For example, when the user is given a hint in the form of the edge completion $E(\bfx)$, one possible choice for the corresponding $\lambda_i(\bfx)$ is given by
\begin{align}\label{eq:aniso_lambda}
    \lambda_i(\bfx) = (1-\lambda_a) \exp \left(-\frac{DT(E(\bfx))}{\tau}\right) + \lambda_a,
\end{align}
where $\lambda_a$ is the ``asymptotic" value of $\lambda_i(\bfx)$, $\tau$ is the corresponding relaxation time and $DT(E(\bfx))$ is the distance transform of the binary edge map $E(\bfx)$.
We provide several practical examples illustrating the advantages of the choice in \eqref{eq:aniso_lambda} in Section \ref{sec:examples}.

\begin{figure}
    \centering
    \includegraphics[width=0.3\textwidth]{example_3/image.png}
    \qquad
    \begin{overpic}[width=0.3\textwidth,tics=5]{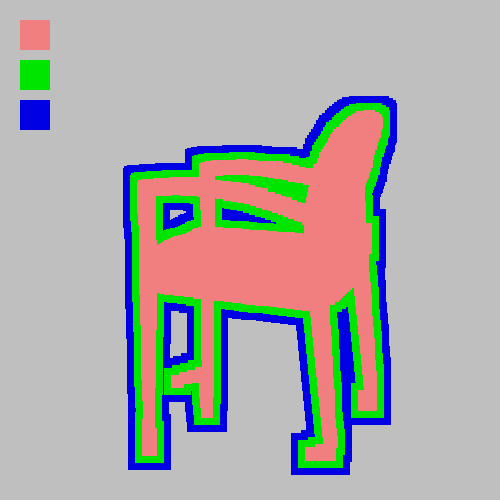}
        \put (12,91) {\scriptsize $\inpdom$}
        \put (12,83) {\scriptsize $\inpdom_*$}
        \put (12,74) {\scriptsize $\tilde{\inpdom}_*$}
        \put (32,87) {\scriptsize $\inpdom \subset \inpdom_* \subset \tilde{\inpdom}_*$}
    \end{overpic}
    \caption{Image of a chair with original and extended inpainting regions $\inpdom \subset \inpdom_* \subset \tilde{\inpdom}_*$.} 
    \label{fig:regionsb}
\end{figure}  

Similarly to \eqref{eq:Arias_model}, we extend the inpainting domain $\mathcal{O}$ so that only information from pixels outside the inpainting region is utilized to determine values of pixels in the inpainting region. To accomplish this, we first define the subset $\mathcal{O}_*$ of $\Omega$ such that for every element $\bfy$ in the complement $\mathcal{O}_*^c = \Omega \backslash \mathcal{O}_*$, the convolution $(g_i * \bfu)(\bfy)$ is determined entirely by pixels outside the inpainting region\footnote{Note that this places limitations on the support of $g_i$ if there is to be any region that is completely determined by pixels outside the inpainting region.}:
\begin{equation}\label{def:Ostart}
    \inpdom_* = \left\{ \bfx \in \Omega: \left(\bfx + \bigcup_i \supp (g_i) \right) \cap \mathcal{O} \neq \emptyset \right\}.
\end{equation}
In addition, we extend $\inpdom_*$ to the subset $\tilde{\inpdom}_*$ of $\Omega$ such that for every element $\bfy$ in the complement $\tilde{\inpdom}^c_* = \Omega \backslash \tilde{\inpdom}_*$, the patch about $\bfy$ is a subset of $\inpdom_*^c$ ({\em cf}.\eqref{eqn:patchdef}):
\begin{equation}\label{def:Otiledstar}
    \tilde{\mathcal{O}}_* = \left\{ \bfx \in \Omega : (\bfx + \mathcal{P}) \cap \mathcal{O}_* \neq \emptyset \right\}.
\end{equation}
See Figure \ref{fig:regionsb} for an illustration of the regions $\inpdom\subset\inpdom_*\subset\tilde{\inpdom}_*$. 
The motivation for defining \eqref{def:Otiledstar} is that for each $\bfy \in \tilde{\inpdom}_*^c$ and $\bfh \in \mathcal{P}$, the quantity $(g_i * \bfu)(\bfy + \bfh)$ is calculated entirely on $\inpdom^c$, the complement of the inpainting region. 
Consequently, restricting $\bfy$ to the set $\tilde{\inpdom}_*^c$ in \eqref{eq:our_dist} ensures that structure at $\bfx$ is only compared to information in the complement of the inpainting region $\inpdom^c$.

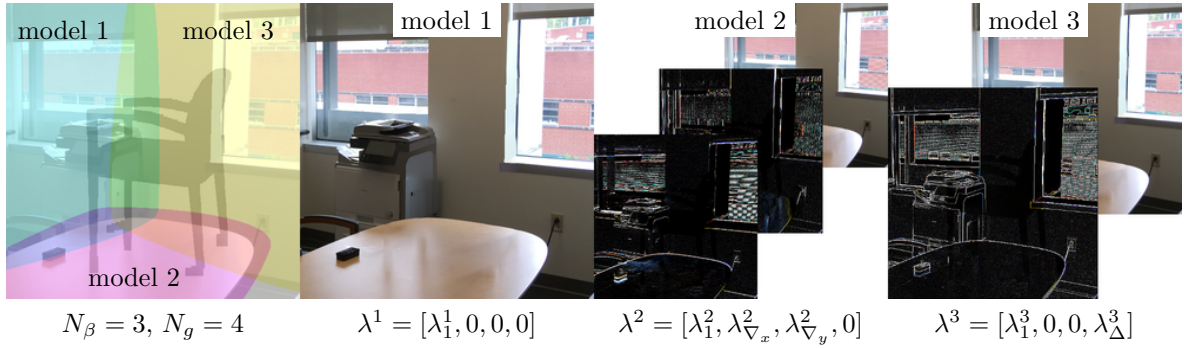
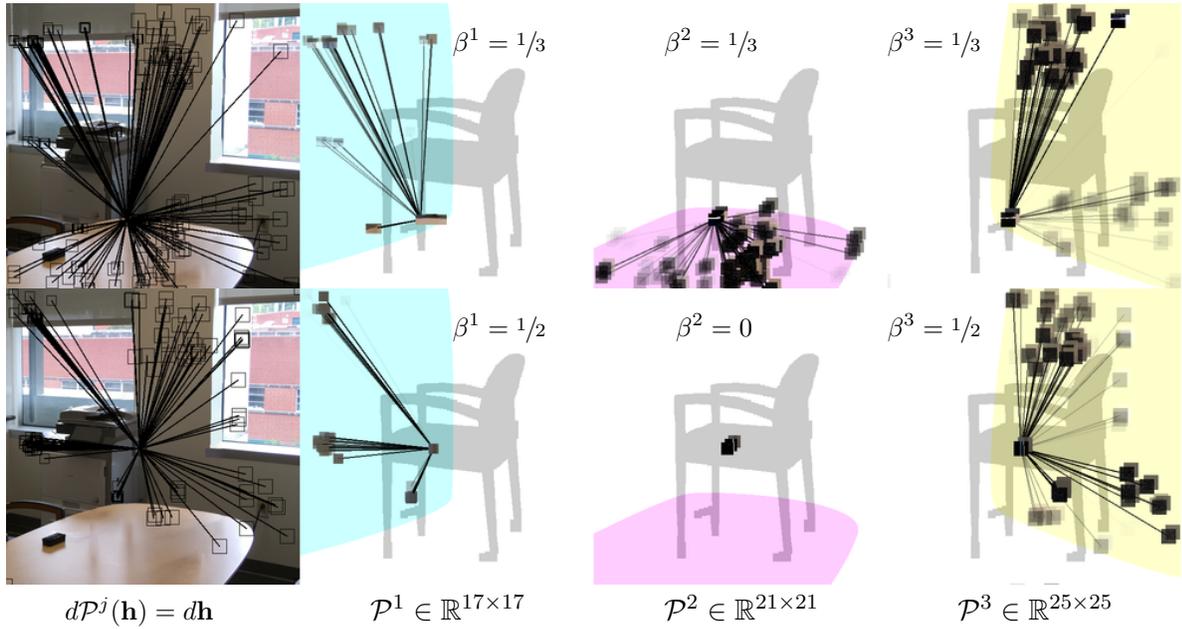
\begin{figure}[t]
    \centering\begin{subfigure}{\textwidth}
        \begin{overpic}[width=\textwidth,tics=5]{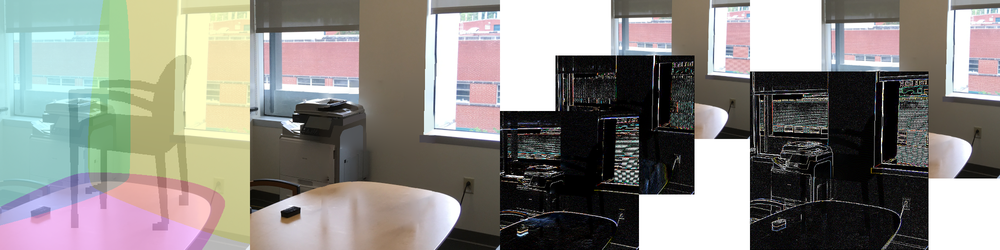} 
            \put (1,22)  {\small model $1$}
            \put (7,1)   {\small model $2$}
            \put (15,22) {\small model $3$}
            \put (0,23){\small
            \begin{tabularx}{\textwidth}{ 
                  >{\centering\arraybackslash}X
                  >{\centering\arraybackslash}X
                  >{\centering\arraybackslash}X
                  >{\centering\arraybackslash}X
                  }
                  & \colorbox{white}{model $1$} & \colorbox{white}{model $2$} & \colorbox{white}{model $3$}
            \end{tabularx}}
            \put (0,-3){\small
            \begin{tabularx}{\textwidth}{ 
                  >{\centering\arraybackslash}X
                  >{\centering\arraybackslash}X
                  >{\centering\arraybackslash}X
                  >{\centering\arraybackslash}X
                  }
                 $N_{\beta}=3$, $N_{g}=4$ & $\lambda^1=[\lambda_1^1,0,0,0]$ & $\lambda^2=[\lambda_1^2,\lambda_{\nabla_x}^2,\lambda_{\nabla_y}^2,0]$ & $\lambda^3=[\lambda_1^3,0,0,\lambda_{\Delta}^3]$
            \end{tabularx}}
        \end{overpic}
        \vspace{0.2em}
        \caption{Three selected models with their corresponding features: 1.~image intensity $(g_1)$, 2.~image intensity with image gradient $(g_1, g_{\nabla_x}, g_{\nabla_y})$, 3.~image instensity with image Laplacian $(g_1, g_{\Delta})$.}
        \label{fig:feature_graphs_a}
    \end{subfigure}
    \begin{subfigure}{\textwidth}
        \begin{overpic}[width=\textwidth,tics=5]{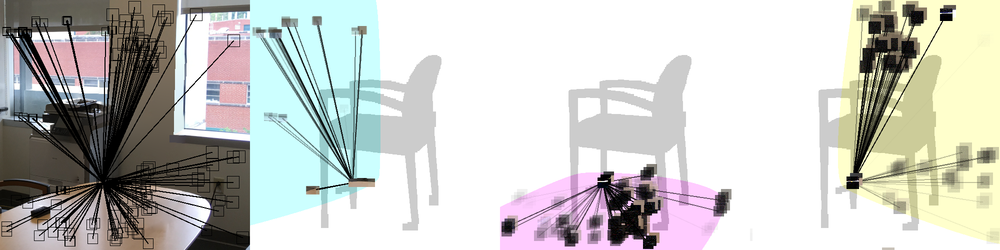} 
            \put (38,21) {\small$\displaystyle\beta^1=\sfrac{1}{3}$}
            \put (56,21) {\small$\displaystyle\beta^2=\sfrac{1}{3}$}
            \put (75,21) {\small$\displaystyle\beta^3=\sfrac{1}{3}$}
        \end{overpic}
        \begin{overpic}[width=\textwidth,tics=5]{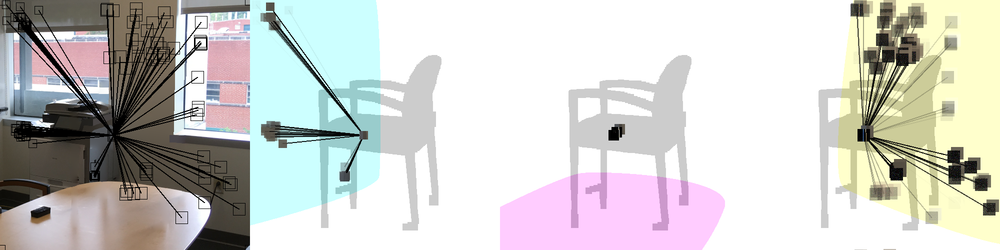} 
            \put (38,21) {\small$\displaystyle\beta^1=\sfrac{1}{2}$}
            \put (57,21) {\small$\displaystyle\beta^2=0$}
            \put (75,21) {\small$\displaystyle\beta^3=\sfrac{1}{2}$}
            \put (5,-3)  {\small$d\patch^j(\bfh) = d\bfh$}
            \put (31,-3) {$\patch^1\in\mathbb{R}^{17\times 17}$}
            \put (56,-3) {$\patch^2\in\mathbb{R}^{21\times 21}$}
            \put (81,-3) {$\patch^3\in\mathbb{R}^{25\times 25}$}
        \end{overpic}
        \vspace{0.2em}
        \caption{Feature graphs of the selected models and their partial contributions for two different pixels.}
        \label{fig:feature_graphs_b}
    \end{subfigure}
    \caption{Possible selection of the features and models. Depicted feature graphs have $50$ nodes, the transparency of patches is used to encode the corresponding weights $w^j(\bfx,\bfy)$.}
    \label{fig:feature_graphs}
\end{figure} 

Now one can define the target functional by the analogy with \eqref{eq:Arias_model} using the modified distance metric in \eqref{eq:our_dist}.
Instead, we extend this definition by introducing $N_{\beta}$ auxiliary feature models with corresponding weights
\begin{align}\label{eq:feature_weights}
    w^{j}(\bfx,\bfy) &= \frac{ \exp \left( - d^j[\bfu,g](\bfx,\bfy)/\sigma^j \right)}{{\int\limits_{\tilde{\inpdom}_*^c} \exp \left( -d^j[\bfu,g](\bfx,\bfy)/\sigma^j \right) d\bfy}},
    \qquad
    j=1,\hdots,N_{\beta}
\end{align} 
and distance metrics
\begin{align*}
    &d^j[\bfu,g](\bfx,\bfy) =
    \sum_{i=1}^{N_{g}} \int\limits_{\patch^j} \lambda_i^j(\bfx+\bfh) \Big| (g_i*\bfu)(\bfx+\bfh) - (g_i*\bfu)(\bfy+\bfh) \Big|^2 d\patch^j(\bfh)
\end{align*}
such that the proposed functional takes the form of an advanced \textit{mixture of $N_\beta$ feature models}
\begin{align}\label{eq:our_model}
    \mathcal{E}[\bfu,w] &:=
    \sum_{j=1}^{N_{\beta}} \int\displaylimits_{\tilde{\inpdom}_*} \beta^j(\bfx) \int\displaylimits_{\tilde{\inpdom}_*^c} w^j(\bfx,\bfy) d^j[\bfu, g](\bfx,\bfy) d\bfy d\bfx,
    \quad \text{s.t. } \sum_{j=1}^{N_{\beta}} \beta^j(\bfx) = 1 \qquad \forall \bfx,
\end{align}
where the coefficients $\beta^j(\bfx)$ define the partial contributions of corresponding models at each~$\bfx$.
Figure~\ref{fig:feature_graphs} illustrates this concept for the case of $N_g=4$ selected features and $N_{\beta}=3$ models.
For this illustration, we used the uniform patch weights $d\patch^j(\bfh) = d\bfh$ and different patch sizes $\patch^j$ for each model.
In \eqref{eq:feature_weights}-\eqref{eq:our_model}, all $N_g$ features $g_i*u$ are accessible to all $N_{\beta}$ models, and the coefficients $\lambda_i^j$ are then used to define the active features in each model.
For instance, consider Figure~\ref{fig:feature_graphs_a}, where we illustrate an example with $N_\beta=3$ and $N_g=4$. 
The first model's coefficients are $\lambda^1=\left[\lambda_1^1,0,0,0\right]$ which limits the active features to the image intensity, the second model's coefficients are $\lambda_2 = \left[\lambda_1^2, \lambda_{\nabla_x}^2, \lambda_{\nabla_y}^2,0 \right]$ which selects image intensity and the image gradient as active features, and the third model's coefficients are $\lambda^3 = \left[\lambda_1^2,0,0,\lambda_\Delta^3 \right]$ which selects the image intensity and image Laplacian as active features. 
Additionally, Figure \ref{fig:feature_graphs_b} shows the feature graphs and their coefficients $\beta^j(\bfx)$ for two different pixels $\bfx\in\inpdom$.
One can clearly observe the variation in the contribution of the selected feature models to the value of the energy functional in different parts of the image.
Note that in the particular case of scalar~$\beta^j$, each model contributes globally to the energy functional in \eqref{eq:our_model} which is then given by the simple weighted sum of the corresponding functionals 
\begin{align}\label{eq:mixture_model}
    \mathcal{E}[\bfu,w]
    :=\sum_{j=1}^{N_{\beta}} \beta^j \mathcal{E}^j[\bfu,w]
    \qquad\text{with}\qquad
    \mathcal{E}^j[\bfu,w]:= \int\limits_{\tilde{\inpdom}_*} \int\limits_{\tilde{\inpdom}_*^c} w^j(\bfx,\bfy) d^j[\bfu, g](\bfx,\bfy) d\bfy d\bfx.
\end{align}
Additionally note that the minimizer of \eqref{eq:mixture_model} is not a weighted sum of $N_{\beta}$ minimizers but rather a solution representing the optimal balance between $N_{\beta}$ models.
We illustrate the case of scalar and more general $\beta^j$ in Section \ref{sec:examples}. Specifically, see Examples 2 and 4 as well as the corresponding Figures \ref{fig:ex_2_result} and \ref{fig:ex_4_result}.

The described concept of the feature models grants the user more precise control over the desired local behavior of the algorithm in the specified parts of the image.
It should be noted that several approaches to add an adaptive component to exemplar-based algorithms have appeared in the literature before including techniques for the locally optimal selection of patch shapes \cite{Guillemot2013} and the design of affine-invariant metrics \cite{fedorov2016}.
Our approach differs in that 1) we allow for the selection and combination of the features of interest, and 2) our method is not a postprocessing of the solution but rather a mixture of models at a more fundamental level of the variational problem formulation.

In order to minimize the functional \eqref{eq:our_model}, we employ an iterative approach of alternating between the updating of $\bfu$ and $w(\bfx,\bfy)$. This approach is inspired by algorithms appearing in \cite{Peyre2011} and \cite{Arias2011} for minimizing \eqref{eqn:graphbasedfunctional} and \eqref{eq:Arias_model} respectively. Before we present our algorithm, we first explore the weight and image update steps individually. We begin with the weight updating step.

\begin{algorithm}[t]
    \caption{Feature driven exemplar inpainting}
    \begin{algorithmic}[1]
    	\Statex
    	\setcounter{ALG@line}{0}
        \Function{UpdateWeights}{$u$}
    	        \For{$j\in1\hdots N_\beta$}
    	            \For{$\;\bfx\in\tilde{\mathcal{O}}_*,\;\bfy\in\tilde{\mathcal{O}}_*^c$}
    	                \State Evaluate the patch distance 
    	                $$d^j[\bfu,g](\bfx,\bfy)\leftarrow\sum_{i=1}^{N_{g}} \int\displaylimits_{\patch^j} \lambda_i^j(\bfx+\bfh) \Big| (g_i*\bfu)(\bfx+\bfh) - (g_i*\bfu)(\bfy+\bfh) \Big|^2 d\patch^j(\bfh)$$
    	                \State Evaluate the weights $$w^j(x,y) \leftarrow \frac{\exp \left( -d^j[u](x,y)/\sigma^j \right)}{\int\limits_{\tilde{\mathcal{O}}_*^c} \exp \left( -d^j[u](x,y)/\sigma^j \right) dy}$$
    	            \EndFor
    	        \EndFor
    	    \Return $w$
        \EndFunction
        \Statex
        \setcounter{ALG@line}{0}
        \Function{UpdateImage}{$w$}
            \For{$\;\bfx\in\mathcal{O}_*$}
                \State Evaluate the auxiliary functions 
                $$k^j(\bfx) = \int\displaylimits_{\inpdom_*^c} \int\displaylimits_{\patch^j} \beta^j(\bfx-\bfh) w^j(\bfx-\bfh,\hat{\bfx}-\bfh) d\patch^j(\bfh) d\hat{\bfx}$$
                $$f_i^j(\bfx) = \int\displaylimits_{\inpdom_*^c} \int\displaylimits_{\patch^j} \beta^j(\bfx-\bfh) w^j(\bfx-\bfh,\hat{\bfx}-\bfh) (g_i * \bfu^k) (\hat{\bfx}) d\patch^j(\bfh) d\hat{\bfx}$$
            \EndFor
            \State Solve the boundary-value problem $$\sum_{j=1}^{N_\beta} \sum_{i=1}^{N_g} \Big[ \overline{g}_i*\big(\lambda_i^j k^j (g_i*\bfu) \big) \Big] (\bfx)
    = \sum_{j=1}^{N_\beta} \sum_{i=1}^{N_g} \Big[ \overline{g}_i * (\lambda_i^j f_i^j) \Big] (\bfx)$$ 
        \Return $\bfu$
        \EndFunction
    \end{algorithmic}
    \label{alg:our_algorithm}
\end{algorithm}

\subsection{Updating weights}\label{sec:weightupdate}
When the image $\bfu$ is held fixed, updating the weights is relatively simple. For convenience, in the remainder of this work we suppose $\sigma^j\to 0$ in \eqref{eq:feature_weights}, in which case the weights tend to delta distributions
\begin{align}\label{eq:delta_weights}
    w^j(\bfx,\bfy) \to \delta \big(d^j[\bfu,g](\bfx,\bfy)\big),
\end{align}
or more precisely
\begin{align*}
    w^j(\bfx,\bfy) = 
    \begin{cases}
        1, & \bfy = \varphi^j(\bfx),
        \\
        0, & otherwise,
    \end{cases}
\end{align*}
where $\varphi^j(\bfx)$ is the nearest neighbor field of $\tilde{\inpdom}_*$, i.e.,
\begin{align*}
    \varphi^j(\bfx) = \argmin_{\bfy \in\tilde{\inpdom}_*^c}d^j[\bfu,g](\bfx,\bfy), 
    \qquad
    \bfx \in \tilde{\inpdom}_*.
\end{align*}

\subsection{Updating image}

With the assumption that $\sigma^j \rightarrow 0$ in \eqref{eq:feature_weights}, the Euler-Lagrange equation of \eqref{eq:our_model} is given by
\begin{align}
    \label{eqn:our_bdyproblem}
    \sum_{j=1}^{N_\beta} \sum_{i=1}^{N_g} \Big[ \overline{g}_i*\big(\lambda_i^j k^j (g_i*\bfu) \big) \Big] (\bfx)
    = \sum_{j=1}^{N_\beta} \sum_{i=1}^{N_g} \Big[ \overline{g}_i * (\lambda_i^j f_i^j) \Big] (\bfx), \qquad \bfx \in \inpdom,
\end{align}
where
\begin{align}
    m^j(\bfz,\hat{\bfz}) &:= \int\limits_{\patch_j} \beta^j(\bfz-\bfh) w^j(\bfz-\bfh,\hat{\bfz}-\bfh) d\patch^j(\bfh), \label{def:shorthandforimageupdate1} \\
    k^j(\bfz) &:= \int\limits_{\inpdom_*^c} m^j(\bfz,\hat{\bfz}) d\hat{\bfz}, \label{def:shorthandforimageupdate2} \\
    f_i^j(\bfz) &:= \int\limits_{\inpdom_*^c} m^j(\bfz,\hat{\bfz}) (g_i * \bfu) (\hat{\bfz}) d\hat{\bfz}. \label{def:shorthandforimageupdate3}
\end{align}
The derivation of \eqref{eqn:our_bdyproblem} is presented in Appendix \ref{appendix:algorithm}.
The well-posedness of the boundary value problem \eqref{eqn:our_bdyproblem} depends on the properties of the convolution operators $g_i$. 
Without loss of generality, we suppose $g_1(t) = \delta(t)$ so that this term acts as a classical Tikhonov regularizer. 
When $g_1(t)$ is the only selected filter for each model, the problem in \eqref{eqn:our_bdyproblem} admits a simple analytical solution 
\begin{align}\label{eq:our_explicit_solution}
    \bfu(\bfx) = \sum_{j=1}^{N_{\beta}} s^j(\bfx) \int\displaylimits_{\tilde{\inpdom}^c} &\int\displaylimits_{\patch^j} \beta^j(\bfx-\bfh)
    w^j(\bfx-\bfh,\hat{\bfx}-\bfh) d\patch^j(\bfh) \bfu(\hat{\bfx})  d\hat{\bfx},
\end{align}
where $$s^j(\bfx) =  \frac{\lambda_1^j(\bfx)}{ \sum\limits_{l=1}^{N_{\beta}} \lambda_1^l(\bfx) \int\limits_{\patch^l} \beta^l(\bfx-\bfh) d\patch^l(\bfh)}.$$
With the weights in \eqref{eq:delta_weights}, the above expression can be simplified to
\begin{align*}
    &\bfu(\bfx) = 
    \left. \frac{\sum\limits_{j=1}^{N_{\beta}} \lambda_1^j(\bfx) \int\limits_{\patch^j} \beta^j(\bfx-\bfh) \bfu\big(\varphi^j(\bfx-\bfh)+\bfh\big) d\patch^j(\bfh)}{
    \sum\limits_{l=1}^{N_{\beta}} \lambda_1^l(\bfx) \int\limits_{\patch^l} \beta^l(\bfx-\bfh) d\patch^l(\bfh)} \right.
\end{align*}
and in case of scalar $\beta^j$, it further simplifies to
\begin{align*}
    &\bfu(\bfx) = 
    \sum_{j=1}^{N_{\beta}} s^j \int\displaylimits_{\patch^j} \bfu\big(\varphi^j(\bfx-\bfh)+\bfh\big) d\patch^j(\bfh)
    \qquad\text{with}\quad
    s^j=\frac{\beta^j \lambda_1^j(\bfx)}{\sum\limits_{l=1}^{N_{\beta}} \beta^l \lambda_1^l(\bfx)}.
\end{align*}

One can view this and the more general solution in \eqref{eq:our_explicit_solution} as a result of applying to the image a cleverly designed nonlinear filter. 
In this regard, it can also be considered as a generalization of the patch nonlocal means filter proposed in~\cite{Arias2011}. 

The discretization of the system in \eqref{eqn:our_bdyproblem} is obtained trivially by replacing all integrals with sums and all continuous convolutions with their discrete analogs. 
Algorithm \ref{alg:our_algorithm} provides the details of the proposed inpainting technique.

\begin{figure}[t]
    \centering
    \stackinset{c}{0pt}{b}{-15pt}{\small Original image $u$}{\includegraphics[width=0.3\textwidth]{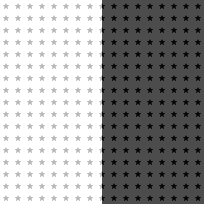}}
    \qquad
    \stackinset{c}{0pt}{b}{-30pt}{\small \Centerstack{Inpainting domain (red) \\ and available exemplars (green)}}{\includegraphics[width=0.3\textwidth]{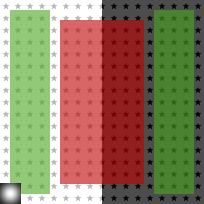}}
    \caption{Setup of Example 1. A $21\times 21$ Gaussian patch weight with a standard deviation $\sigma=10$ is shown in the left bottom corner.}
    \label{fig:ex_1_setup}
\end{figure}

\section{Numerical examples}\label{sec:examples}

In all examples below, we used the custom implementation of the PatchMatch method \cite{Barnes:2009:PAR} for the evaluation of the nearest neighbor fields.
The source code, including the library and all examples, can be found at \url{https://github.com/vreshniak/feature-driven-exemplar-inpainting}.

\paragraph{Example 1}

As a first example, consider the simple synthetic image in Figure~\ref{fig:ex_1_setup}.
The size of the image is $200$ by $200$ pixels and the inpainitng domain 
is chosen such that the available exemplar regions (shown in green) are disjointly located in the parts of the image with the opposite intensity. 

We employ the following kernels and their adjoints 
\begin{subequations}
    \begin{alignat}{3}
        \label{eq:id_ker}
        &g_1 =
        \left[\begin{array}{c}
    	    1
    	\end{array}\right],
    	\qquad
    	&&\overline{g}_1 =
        \left[\begin{array}{c}
    	    1
    	\end{array}\right],
    	\\
    	\label{eq:dx_ker}
    	&g_{\nabla_x} = 
        \left[\begin{array}{rrr}
        	  &    &   \\
        	0 & -1 & 1 \\
        	  &    &
    	\end{array}\right],
    	\qquad
    	&&\overline{g}_{\nabla_x} =
    	\left[\begin{array}{rrr}
        	   &   &   \\
        	-1 & 1 & 0 \\
        	   &   &
    	\end{array}\right],
    	\\
    	\label{eq:dy_ker}
    	&g_{\nabla_y} = 
    	\left[\begin{array}{rrr}
        	&  0 & \\
        	& -1 & \\
        	&  1 &
    	\end{array}\right],
    	\qquad
    	&&\overline{g}_{\nabla_y} =
    	\left[\begin{array}{rrr}
        	& -1 & \\
        	&  1 & \\
        	&  0 &
    	\end{array}\right],
    	\\
    	\label{eq:laplacian_ker}
    	&g_{\Delta} = 
        \left[\begin{array}{rrr}
        	  &  1 &   \\
        	1 & -4 & 1 \\
        	  &  1 & 
    	\end{array}\right],
    	\qquad
    	&&\overline{g}_{\Delta} =
        \left[\begin{array}{rrr}
        	  &  1 &   \\
        	1 & -4 & 1 \\
        	  &  1 & 
    	\end{array}\right].
    \end{alignat}
\end{subequations}
One can recognize $g_1$ as the identity map while $(g_{\nabla_x},g_{\nabla_y})$ and $g_{\Delta}$ correspond to the finite difference stencils of the image gradient and image Laplacian respectively.

For the given patch $\patch$, we consider patch weights as probability measures of the form
\begin{align}\label{eq:ex_1_patch}
    d\patch(\bfh) 
    = \frac{\exp \left( -\norm{\bfh}{}^2 /\sigma^2 \right)}{\int\limits_{\patch{}} \exp \left( -\norm{\bfp}{}^2/\sigma^2 \right) d\bfp} d\bfh
\end{align}
such that $\int\limits_{\patch} d\patch(\bfh) = 1$.
We also consider a single model so that $N_{\beta}=1$ and $\beta(\bfx):=1$ in \eqref{eq:our_model} and set all $\lambda_i^1:=\lambda_i$ to be constants.
With this choice, we get
\begin{align*}
    &k(\bfz) = \int\limits_{\patch} d\patch(\bfh) = 1,
    \quad \forall \bfz
\end{align*}
and the boundary value problem \eqref{eqn:our_bdyproblem} simplifies to
\begin{align*} 
    \sum_{i=1}^{N_g} \lambda_i \Big(\overline{g}_i*g_i*\bfu \Big)(\bfx) 
    = \sum_{i=1}^{N_g} \lambda_i \Big(\overline{g}_i * f_{i} \Big)(\bfx), &\quad \bfx \in \inpdom,
    \\
    \bfu(\bfx) = \hat{\bfu}(\bfx), &\quad \bfx \in \inpdom^c.
\end{align*}

\begin{figure}[t]
    \centering
    \begin{subfigure}[t]{.24\linewidth}
        \centering
        \includegraphics[width=0.98\textwidth]{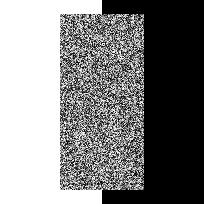}
        \caption{Initialization.}
        \label{fig:ex_1_step_a}
    \end{subfigure}
    \begin{subfigure}[t]{.24\linewidth}
        \centering
        \includegraphics[width=0.98\textwidth]{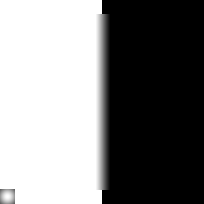}
        \caption{Inpainting using $g_1$.}
        \label{fig:ex_1_step_b}
    \end{subfigure}
    \begin{subfigure}[t]{.24\linewidth}
        \centering
        \includegraphics[width=0.98\textwidth]{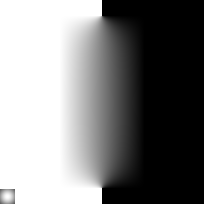}
        \caption{Inpainting using $g_{\nabla}$.}
        \label{fig:ex_1_step_c}
    \end{subfigure}
    \begin{subfigure}[t]{.24\linewidth}
        \centering
        \includegraphics[width=0.98\textwidth]{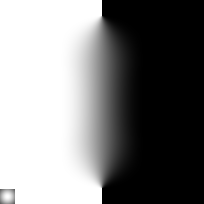}
        \caption{Inpainting using $g_{\Delta}$.}
        \label{fig:ex_1_step_d}
    \end{subfigure}
    \\[1em]
    \begin{subfigure}[t]{.24\linewidth}
        \centering
        \includegraphics[width=0.98\textwidth]{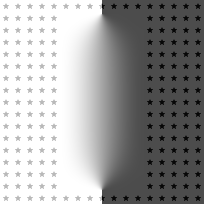}
        \caption{Initialization.}
        \label{fig:ex_1_PDE_a}
    \end{subfigure}
    \begin{subfigure}[t]{.24\linewidth}
        \centering
        \includegraphics[width=0.98\textwidth]{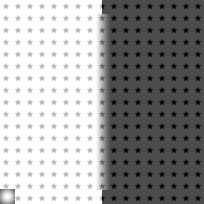}
        \caption{Inpainting using $g_1$.}
        \label{fig:ex_1_PDE_b}
    \end{subfigure}
    \begin{subfigure}[t]{.24\linewidth}
        \centering
        \includegraphics[width=0.98\textwidth]{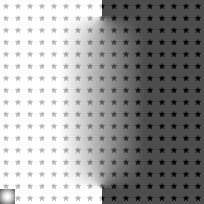}
        \caption{Inpainting using $g_{\nabla}$.}
        \label{fig:ex_1_PDE_c}
    \end{subfigure}
    \begin{subfigure}[t]{.24\linewidth}
        \centering
        \includegraphics[width=0.98\textwidth]{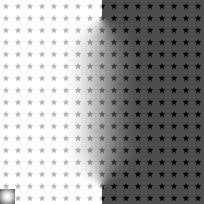}
        \caption{Inpainting using $g_{\Delta}$.}
        \label{fig:ex_1_PDE_d}
    \end{subfigure}
    \caption{Ipainting of the step image in Example 1 using $15\times 15$ patch  $\patch$ with $\sigma=10$.}
    \label{fig:ex_1_PDE}
\end{figure}

To better illustrate the impact of the choice of the convolutional filters $g_i$ on the result of the inpainting procedure, we start with the trivial step image in Figure~\ref{fig:ex_1_step_a} initialized with random noise.
We choose a $15\times 15$ patch with $\sigma=10$ and consider three different cases.
In the first case, we set $\lambda_1=1$ and $\lambda_i=0$, $i\neq 1$; this leads to the following image update step 
\begin{align*}
    \bfu(\bfx) = 
    \int\displaylimits_{\patch} \bfu\big(\varphi(\bfx-\bfh)+\bfh\big) d\patch(\bfh)
\end{align*}
which simply assigns the value of each pixel $\bfx$ based on the ``votes" of the nearest neighbors of the pixels in the patch around $\bfx$.
The result is presented in Figure \ref{fig:ex_1_step_b}.
One can see that the algorithm is capable of the exact recovery of the constant image intensity in the regions where all pixels in the patch around $\bfx$ have nearest neighbors in either left or right parts of the image.
The smooth transition between constant regions happens in the band whose width coincides with the size of the patch.

For the second case we set $\lambda_{\nabla_x}=\lambda_{\nabla_y}=1$ and $\lambda_i=0$ otherwise; this choice gives
\begin{align*}
    &f_{\nabla_x}(\bfz) = \int\displaylimits_{\patch} (g_{\nabla_x} * \bfu) \big(\varphi(\bfz-\bfh)+\bfh\big) d\patch(\bfh) = 0,
    \\
    &f_{\nabla_y}(\bfz) = \int\displaylimits_{\patch} (g_{\nabla_y} * \bfu) \big(\varphi(\bfz-\bfh)+\bfh\big) d\patch(\bfh) = 0
\end{align*}
since $\bfu$ is constant in $\tilde{\inpdom}_*^c$.
As a result, we get the following homogeneous boundary value problem
\begin{align*}
    \Big((\overline{g}_{\nabla_x}*g_{\nabla_x} + \overline{g}_{\nabla_y}*g_{\nabla_y})*\bfu\Big)(\bfx)
    = 0, &\quad \bfx \in \inpdom,
    \\
    \bfu(\bfx) = \hat{\bfu}(\bfx), &\quad \bfx \in \inpdom^c.
\end{align*}
The direct substitution shows
\[
\setlength{\arraycolsep}{3pt}
    \overline{g}_{\nabla_x}*g_{\nabla_x} + \overline{g}_{\nabla_y}*g_{\nabla_y} 
    =
	\left[\begin{array}{rrr}
    	   &   &   \\
    	-1 & 1 & 0 \\
    	   &   &
	\end{array}\right]
	*
    \left[\begin{array}{rrr}
    	  &    &   \\
    	0 & -1 & 1 \\
    	  &    &
	\end{array}\right]
	+
    \left[\begin{array}{rrr}
    	& -1 & \\
    	&  1 & \\
    	&  0 &
	\end{array}\right]
	*
    \left[\begin{array}{rrr}
    	&  0 & \\
    	& -1 & \\
    	&  1 &
	\end{array}\right]
	=
    \left[\begin{array}{rrr}
    	  &  1 &   \\
    	1 & -4 & 1 \\
    	  &  1 & 
	\end{array}\right]
\]
which is nothing but the finite difference approximation of the Dirichlet problem for the Laplace equation.
The produced result is shown in Figure \ref{fig:ex_1_step_c} and is identical to the standard harmonic image extension \cite{Chan2005image}.

Analogously, the standard local biharmonic inpainting corresponds to the choice of $g_{\Delta}$ as the convolutional kernel which gives
\begin{align*}
    g_{\Delta}*\overline{g}_{\Delta} = 
    \left[\begin{array}{rrr}
    	  &  1 &   \\
    	1 & -4 & 1 \\
    	  &  1 & 
	\end{array}\right]
    *
    \left[\begin{array}{rrr}
    	  &  1 &   \\
    	1 & -4 & 1 \\
    	  &  1 & 
	\end{array}\right]
	=
    \left[\begin{array}{rrrrr}
	  &    &  1 &    &   \\
	  &  2 & -8 &  2 &   \\
	1 & -8 & 20 & -8 & 1 \\
	  &  2 & -8 &  2 &   \\
	  &    &  1 &    &   \\
	\end{array}\right],
\end{align*}
i.e., the finite difference approximation of the biharmonic differential operator.
Figure \ref{fig:ex_1_step_d} presents the corresponding result.

It is clear that the results of the harmonic and biharmonic inpainting schemes are completely specified by the known values of the image at the boundary of the inpainting domain and do not depend on the provided initialization as opposed to the nonlocal methods. 
The benefits of the nonlocal approach become apparent when considering textured images. 
One can see from Figure~\ref{fig:ex_1_PDE_a} that the local scheme fails to recover the original image structure while the same operator equation with the nonlocal forcing term $\sum_i \lambda_i (\overline{g}_i*f_i)(\bfx)$ successfully recovers texture.

It is worth noting that the inpainting schemes based on the convolutions $g_1$ and $g_{\nabla}=(g_{\nabla_x},g_{\nabla_y})$ can be viewed as particular implementations of the \textbf{\textit{patch nonlocal means}} and the \textbf{\textit{patch nonlocal Poisson}} methods proposed earlier in \cite{Arias2011}.
The proposed framework is hence more general and extends capabilities of the algorithm in~\cite{Arias2011}.
The presented case of the biharmonic inpainting based on the convolution $g_{\Delta}$ is the most obvious such extension.
The nonlocal variants of the above operators can be also considered.
For example, the nonlocal gradient and divergence operators are defined as follows \cite{Gilboa2008}

\noindent
\resizebox{\linewidth}{!}{
\begin{minipage}{\linewidth}
\begin{align*}
    \nabla_{\gamma} \bfu(\bfx,\bfy) &= \big( \bfu(\bfy) - \bfu(\bfx) \big) \gamma(\bfx,\bfy),
    \\
    \nabla_{\gamma} \cdot \overrightarrow{\bfv}(\bfx) &= \int\displaylimits_{\mathcal{K}} \big( \overrightarrow{\bfv}(\bfx,\bfx-\bfh) - \overrightarrow{\bfv}(\bfx-\bfh,\bfx) \big) \gamma(\bfh) d\bfh,
\end{align*}
\end{minipage}
}

\noindent
where $\gamma(\bfh):=\gamma(\bfx,\bfx+\bfh)$ is a symmetric translation invariant kernel with support on a patch $\mathcal{K}$ centered at a pixel $\bfx$.
The nonlocal Laplacian operator is then defined as
\begin{align*}
    \nabla_{\gamma^2}^2 \bfu(\bfx) = 2 \int\displaylimits_{\mathcal{K}} \big( \bfu(\bfx-\bfh) - \bfu(\bfx) \big) \gamma^2(\bfh) d\bfh.
\end{align*}
The discrete versions of these operators can be easily written as convolutions.
For example, consider a $3\times 3$ patch $\mathcal{K}$ with $\left\{\bfh_i \right\}_{i=1}^8$ as the eight non-null coordinates in the patch. The corresponding kernels of the nonlocal gradient have the form
\begin{subequations}
    \setlength{\arraycolsep}{3pt}
    \begin{align*}
    	&g_{\gamma 1} = 
    	\left[\begin{array}{rrr}
        	\gamma(\bfh_1)&  & \\
        	& -\gamma(\bfh_1) & \\
        	&   & 0
    	\end{array}\right],
    	\quad
    	g_{\gamma 2} = 
    	\left[\begin{array}{rrr}
        	&  \gamma(\bfh_2) & \\
        	& -\gamma(\bfh_2) & \\
        	&  0 & 
    	\end{array}\right],
    	\quad\hdots, \quad
    	g_{\gamma 8} = 
    	\left[\begin{array}{rrr}
        	0 &   & \\
        	& -\gamma(\bfh_8) & \\
        	&   & \gamma(\bfh_8)
    	\end{array}\right].
    \end{align*}
\end{subequations}
and the kernel of the discrete nonlocal Laplacian is defined respectively as
\begin{align*}
	g_{\gamma^2} = \frac{1}{2} \sum_{i=1}^8 g_{\gamma i} * \overline{g}_{\gamma i} 
	= 
    \left[\begin{array}{*3{>{\displaystyle}c}}
    	\gamma^2(\bfh_1) & \gamma^2(\bfh_2) & \gamma^2(\bfh_3) \\[1em]
    	\gamma^2(\bfh_4) & -\sum_{j=1}^8\gamma^2(\bfh_j) & \gamma^2(\bfh_5) \\[1.5em]
    	\gamma^2(\bfh_6) & \gamma^2(\bfh_7) & \gamma^2(\bfh_8)
	\end{array}\right].
\end{align*}

The choice of nonlocal kernels instead of local ones might provide new insights into the design of exemplar-based algorithms.
While the idea of embedding images into higher dimensional space of patches allows for the exploration of self-similarity of their local properties globally within the image, nonlocal operators can provide a self-contained similarity measure which makes such embeddings unnecessary.
For example, a nonlocal gradient $g_{\gamma}$ measures variation of the weighted image intensity in a neighbourhood of each pixel similarly to the original patch distance metric \eqref{eq:our_dist}.
Indeed, by taking $d\patch(\bfh)=\gamma^2(\bfh) d\bfh$, $\bfh\in\patch$,
the discrete version of \eqref{eq:our_dist} corresponding to the nonlocal means method reads as
\begin{align*}
    &d[\bfu](\bfx,\bfy) = \sum_{\bfh} \Big| \bfu(\bfx+\bfh) - \bfu(\bfy+\bfh) \Big|^2 \gamma^2(\bfh),
\end{align*}
while for $\bfh\in\mathcal{K}$ we can write the distance metric for the nonlocal gradient as
\begin{align*}
    d[\bfu](\bfx,\bfy) 
    = \sum_{\bfh} \Big| \bfu(\bfx+\bfh) - \bfu(\bfy+\bfh) + \bfu(\bfy) - \bfu(\bfx) \Big|^2 \gamma^2(\bfh).
\end{align*}
Obviously, when $\bfu(\bfx)=\bfu(\bfy)$, these metrics are equivalent. 

\begin{figure}[t]
    \stackinset{c}{0pt}{b}{-15pt}{\small $s=-1$,   $8$   iterations.}{\includegraphics[width=0.24\textwidth]{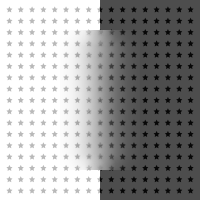}}
    \stackinset{c}{0pt}{b}{-15pt}{\small $s=-0.5$, $10$  iterations.}{\includegraphics[width=0.24\textwidth]{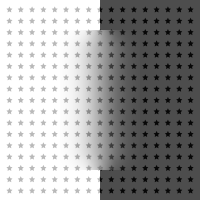}}
    \stackinset{c}{0pt}{b}{-15pt}{\small $s=0.0$,  $20$  iterations.}{\includegraphics[width=0.24\textwidth]{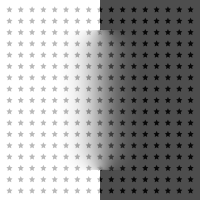}}
    \stackinset{c}{0pt}{b}{-15pt}{\small $s=0.5$,  $200$ iterations.}{\includegraphics[width=0.24\textwidth]{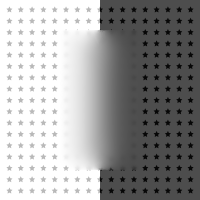}}
    \\[1em]
    \stackinset{c}{0pt}{b}{-15pt}{\small $\sigma=100$, $8$   iterations.}{\includegraphics[width=0.24\textwidth]{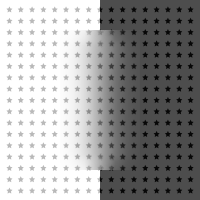}}
    \stackinset{c}{0pt}{b}{-15pt}{\small $\sigma=10$,  $8$   iterations.}{\includegraphics[width=0.24\textwidth]{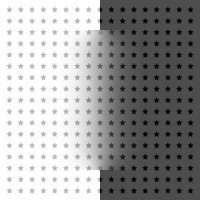}}
    \stackinset{c}{0pt}{b}{-15pt}{\small $\sigma=5$,   $30$  iterations.}{\includegraphics[width=0.24\textwidth]{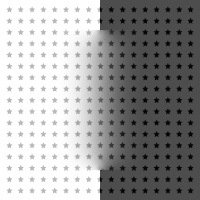}}
    \stackinset{c}{0pt}{b}{-15pt}{\small $\sigma=3$,   $200$ iterations.}{\includegraphics[width=0.24\textwidth]{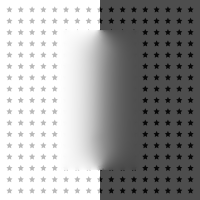}}
    \caption{Inpainting of the step image in Example 1 using $g_{\gamma}$ with $31 \times 31$ kernels (top) $\gamma(\bfh)=\norm{\bfh}{}^{-1-s}$ and (bottom) $\gamma(\bfh)\simeq\exp \left( -\norm{\bfh}{}^2 /\sigma^2 \right)$.}
    \label{fig:ex_1_nonlocal_Laplace}
\end{figure}

When using nonlocal gradients, a patch $\patch$ can be collapsed into a single pixel and the Euler-Lagrange equation \eqref{eqn:our_bdyproblem} takes the simpler form
\begin{align*}
    \sum_{j=1}^{N_\beta} \sum_{i=1}^{N_g} \overline{g}_i*\big(\lambda_i^j \beta^j  (g_i*\bfu) \big) (\bfx)
    = \sum_{j=1}^{N_\beta} \sum_{i=1}^{N_g} \overline{g}_i * (\lambda_i^j f_i^j) (\bfx)
\end{align*}
with $f_i^j(\bfz) = \beta^j(\bfz) \cdot (g_i * \bfu) \big(\varphi^j(\bfz)\big)$.
We will refer to this algorithm as a \textbf{\textit{nonlocal $\gamma$-Poisson}} method.
Despite the similarity of utilized metrics, it has different properties than the nonlocal means method.
For example, it is well known that solutions of the two-dimensional nonlocal Laplace equation
\begin{align*}
    \nabla_{\gamma^2}^2 \bfu(\bfx) = 0
\end{align*}
with 
\begin{align*}
    \gamma(\bfh) = \frac{C_{s}}{ \norm{\bfh}{}^{1+s}},
    \qquad s \in (0,1)
\end{align*}
are in $H^s$, the fractional Sobolev space of order~$s$~\cite{du2013nonlocal}.
Larger values of $s$, which correspond to more localized kernels, lead to smoother interpolation as can be seen from Figure~\ref{fig:ex_1_nonlocal_Laplace}. 
Smaller values of $s$ put more weight on distant pixels instead.
While resulting in less smooth interpolation, such kernels can be useful for filling larger holes in textured images.
For instance, while it took only $8$ iterations for the algorithm with $\gamma\in\mathbb{R}^{31\times 31}$ and $s=-1$ to converge, $20$ iterations were required in the case of $s=0$.

\begin{figure}[t]
    \centering
    	\stackinset{c}{0pt}{b}{-15pt}{\small Possible edge completions $E(\bfx)$ and the corresponding anisotropic coefficients $\lambda(\bfx)$.}{
        \includegraphics[width=0.24\textwidth]{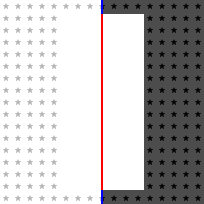}
        \includegraphics[width=0.24\textwidth]{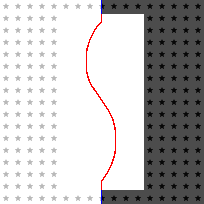}
        \includegraphics[width=0.24\textwidth]{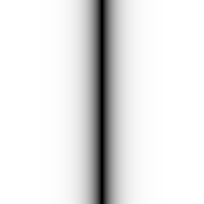}
        \includegraphics[width=0.24\textwidth]{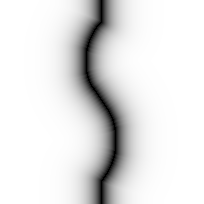}}
    \\[1em]
        \stackinset{c}{0pt}{b}{-15pt}{\small Local vs nonlocal Poisson inpainting.}{
        \includegraphics[width=0.24\textwidth]{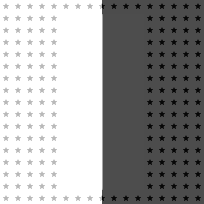}
        \includegraphics[width=0.24\textwidth]{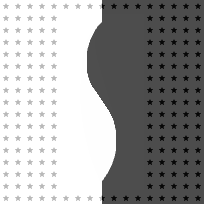}
        \includegraphics[width=0.24\textwidth]{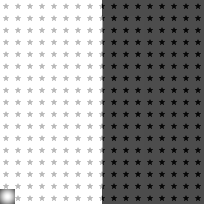}
        \includegraphics[width=0.24\textwidth]{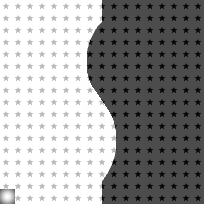}}
    \caption{Inpainting of the image in Example 1 with given edge completion.}
    \label{fig:ex_1_edges}
\end{figure}

Finally, Figure \ref{fig:ex_1_edges} provides an example of the inpainting using manual edge completion $E(\bfx)$ and the anisotropic coefficient $\lambda(\bfx)$ in \eqref{eq:aniso_lambda}.

\paragraph{Example 2}\label{example:2}

\begin{figure}[t]
    \centering
    \stackinset{c}{0pt}{b}{-15pt}{\small Damaged image $u$}{\includegraphics[width=0.32\textwidth]{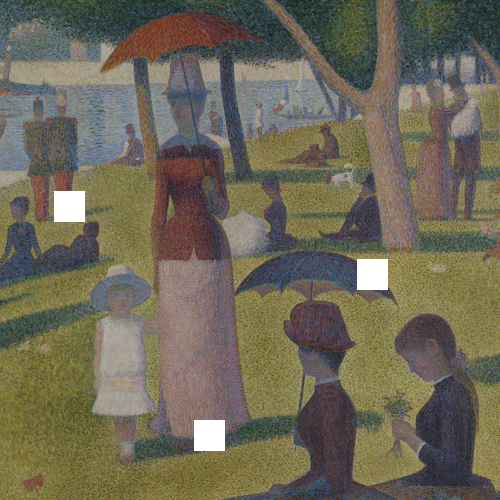}}
    \stackinset{c}{0pt}{b}{-15pt}{\small Image derivative $\nabla_x u$}{\includegraphics[width=0.32\textwidth]{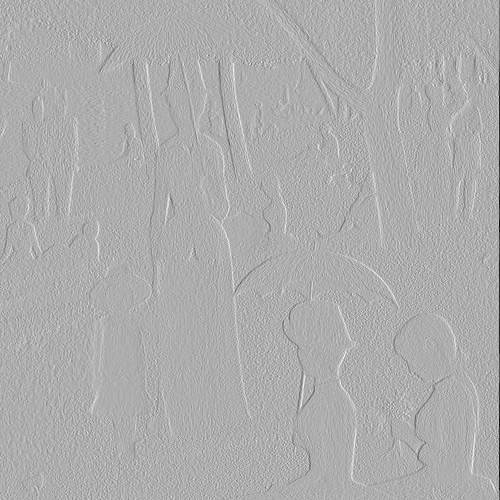}}
    \stackinset{c}{0pt}{b}{-15pt}{\small Gaussian image derivative $\nabla_x^{(5)} u$}{\includegraphics[width=0.32\textwidth]{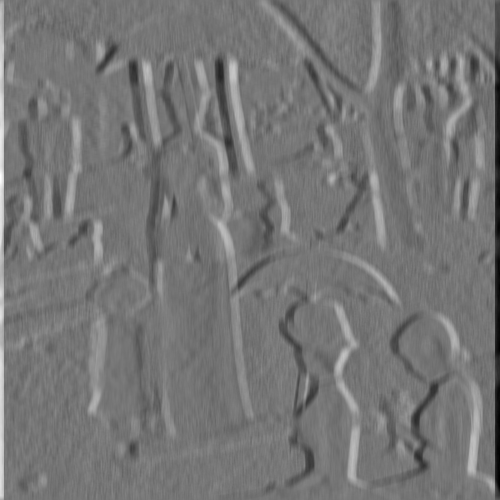}}
    \caption{Setup of Example 2.} 
    \label{fig:ex_2_setup}
\end{figure}

For the second example, consider a fragment of the famous painting by Georges Seurat depicted in Figure~\ref{fig:ex_2_setup}.
The selected fragment is $500 \times 500$ pixels in size and contains both geometric structures and textured regions.

To obtain the results in Figure \ref{fig:ex_2_result}, we used several variants of the model \eqref{eq:our_model} given in Table~\ref{tab:ex_2_models}.
Figure \ref{fig:ex_2_result_a} shows the results of the inpainting procedure for the nonlocal means ($\mathcal{E}^1_\patch$), nonlocal Poisson ($\mathcal{E}^\nabla_\patch$) and nonlocal $\gamma$-Poisson ($\mathcal{E}^\gamma_1$) methods. 
To make the comparison meaningful, we used $15\times 15$ Gaussian patch weights $\patch(\bfh)$ for the nonlocal means and Poisson methods and chose equivalent Gaussian kernels $\gamma(\bfh)$ of the same size for the $\gamma$-Poisson method.
To initialize the algorithms, we downscaled the original image by a factor of ten and then upascaled it to the original size.
This choice provides a rough hint on the large-scale geometric features of the original image and can be considered as a variant of the multi-level initialization strategy~\cite{Modersitzki2009}.

\begin{table}[t]
    \centering
    \def\arraystretch{2.0}
    \setlength{\tabcolsep}{3pt}
    \begin{tabular*}{\textwidth}{rl}
        \hline
        $d^{\alpha,\square}_{1}(\bfx,\bfy)=$      & $\alpha\big| \bfu(\bfx) - \bfu(\bfy) \big|^2 + (1-\alpha) \big\| (g_\square*\bfu)(\bfx) - (g_\square*\bfu)(\bfy) \big\|^2$
        \\
        $d^{\alpha,\square}_{\patch}(\bfx,\bfy)=$ & $\displaystyle{\int\limits_{\patch} d^{\alpha,\square}_{1}(\bfx+\bfh,\bfy+\bfh) d\patch(\bfh)}$
        \\[1em]
        \hline
        $\mathcal{E}^{\alpha,\square}_{\patch}=$ & $\displaystyle{\int\limits_{\tilde{\inpdom}_*} \int\limits_{\tilde{\inpdom}_*^c} w(\bfx,\bfy) d^{\alpha,\square}_\patch(\bfx,\bfy) d\bfy d\bfx}$
        \\
        $\overline{\mathcal{E}^{\beta,\square}_{\patch_1,\patch_2}}=$ & $\beta \mathcal{E}^1_{\patch_1} + (1-\beta) \mathcal{E}^\square_{\patch_2}$
        \\
        $\mathcal{E}^{\square}_{\patch}=$       & $\mathcal{E}^{0,\square}_{\patch}$  \\
        \hline
    \end{tabular*}
    \caption{Models in Example 2. $\square$ is a placeholder for an operator with the kernel~$g_\square$.}
    \label{tab:ex_2_models}
\end{table}

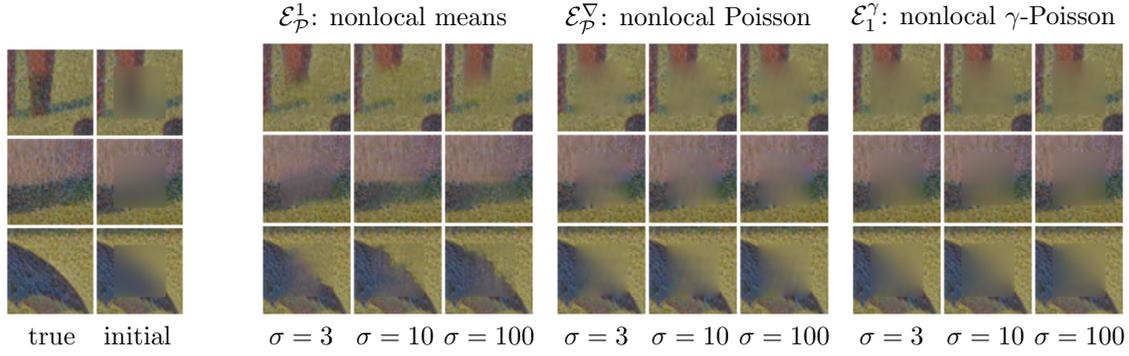
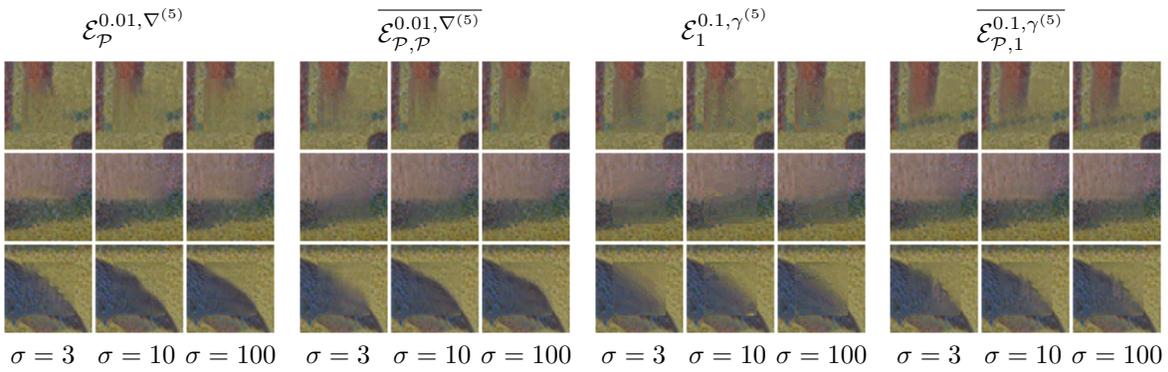
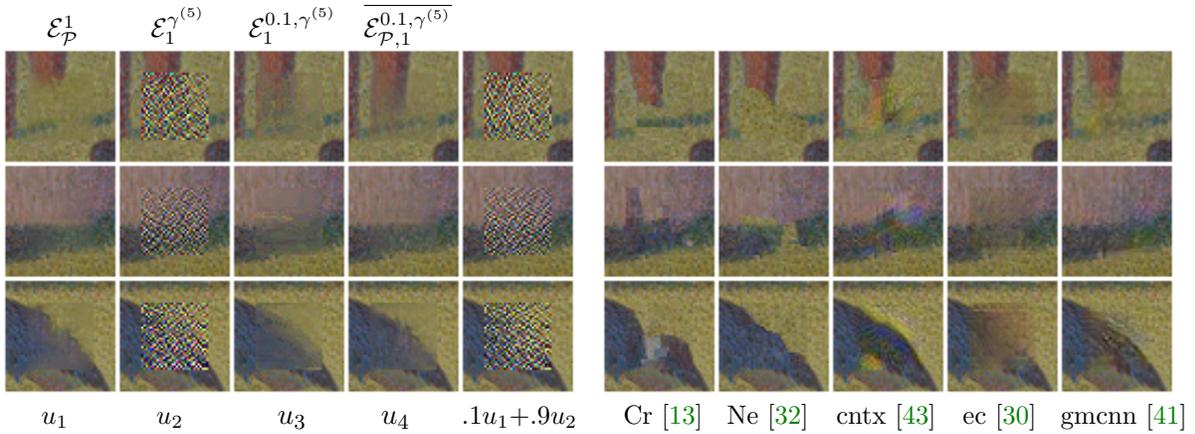
\begin{figure}[!htb]
    \begin{subfigure}[t]{\linewidth}
        \centering 
        \setlength\abovecaptionskip{1.4\baselineskip}
        \stackinset{c}{0pt}{t}{-15pt}{}{
        \begin{overpic}[width=0.23\textwidth,tics=10]{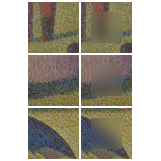} 
            \put (25,-10)   {\small true}
            \put (53,-10)  {\small initial}
        \end{overpic} }
        \stackinset{c}{0pt}{t}{-15pt}{\small $\mathcal{E}^1_\patch$: nonlocal means}{
        \begin{overpic}[width=0.235\textwidth,tics=5]{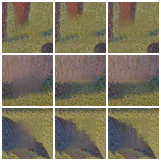} 
            \setlength{\tabcolsep}{0pt}
            \put (-2,-10){\small
            \begin{tabularx}{0.24\textwidth}{ 
                  >{\centering\arraybackslash}X
                  >{\centering\arraybackslash}X
                  >{\centering\arraybackslash}X
                  }
                 $\sigma=3$ & $\sigma=10$ & $\sigma=100$
            \end{tabularx}}
        \end{overpic}}
        \stackinset{c}{0pt}{t}{-15pt}{\small $\mathcal{E}^\nabla_\patch$: nonlocal Poisson}{
        \begin{overpic}[width=0.235\textwidth,tics=5]{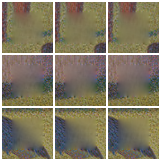} 
            \setlength{\tabcolsep}{0pt}
            \put (-2,-10){\small
            \begin{tabularx}{0.24\textwidth}{ 
                  >{\centering\arraybackslash}X
                  >{\centering\arraybackslash}X
                  >{\centering\arraybackslash}X
                  }
                 $\sigma=3$ & $\sigma=10$ & $\sigma=100$
            \end{tabularx}}
        \end{overpic}}
        \stackinset{c}{0pt}{t}{-15pt}{\small $\mathcal{E}^\gamma_1$: nonlocal $\gamma$-Poisson}{
        \begin{overpic}[width=0.235\textwidth,tics=5]{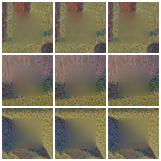} 
            \setlength{\tabcolsep}{0pt}
            \put (-2,-10){\small
            \begin{tabularx}{0.24\textwidth}{ 
                  >{\centering\arraybackslash}X
                  >{\centering\arraybackslash}X
                  >{\centering\arraybackslash}X
                  }
                 $\sigma=3$ & $\sigma=10$ & $\sigma=100$
            \end{tabularx}}
        \end{overpic}}
        \caption{Inpainting with the nonlocal means, nonlocal Poisson and nonlocal $\gamma$-Poisson models using three different Gaussian weights ($\sigma=3,10,100$) for the $15\times 15$ patches $\patch$ and kernels~$\gamma(\bfh)$.} 
        \label{fig:ex_2_result_a}
    \end{subfigure}
    \begin{subfigure}[t]{\linewidth}
        \centering 
        \setlength\abovecaptionskip{1.4\baselineskip}
        \setlength\belowcaptionskip{2\baselineskip}
        \stackinset{c}{0pt}{t}{-20pt}{\small$\mathcal{E}^{0.01,\nabla^{(5)}}_\patch$}{ 
        \begin{overpic}[width=0.235\textwidth,tics=5]{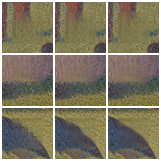} 
            \setlength{\tabcolsep}{0pt}
            \put (-2,-10){\small
            \begin{tabularx}{0.24\textwidth}{ 
                  >{\centering\arraybackslash}X
                  >{\centering\arraybackslash}X
                  >{\centering\arraybackslash}X
                  }
                 $\sigma=3$ & $\sigma=10$ & $\sigma=100$
            \end{tabularx}}
        \end{overpic}}
        \stackinset{c}{0pt}{t}{-20pt}{\small$\overline{\mathcal{E}^{0.01,\nabla^{(5)}}_{\patch,\patch}}$}{ 
        \begin{overpic}[width=0.235\textwidth,tics=5]{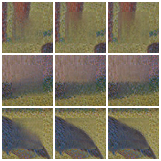} 
            \setlength{\tabcolsep}{0pt}
            \put (-2,-10){\small
            \begin{tabularx}{0.24\textwidth}{ 
                  >{\centering\arraybackslash}X
                  >{\centering\arraybackslash}X
                  >{\centering\arraybackslash}X
                  }
                 $\sigma=3$ & $\sigma=10$ & $\sigma=100$
            \end{tabularx}}
        \end{overpic}}
        \stackinset{c}{0pt}{t}{-20pt}{\small$\mathcal{E}^{0.1,\gamma^{(5)}}_1$}{ 
        \begin{overpic}[width=0.235\textwidth,tics=5]{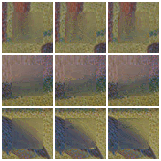} 
            \setlength{\tabcolsep}{0pt}
            \put (-2,-10){\small
            \begin{tabularx}{0.24\textwidth}{ 
                  >{\centering\arraybackslash}X
                  >{\centering\arraybackslash}X
                  >{\centering\arraybackslash}X
                  }
                 $\sigma=3$ & $\sigma=10$ & $\sigma=100$
            \end{tabularx}}
        \end{overpic}}
        \stackinset{c}{0pt}{t}{-20pt}{\small$\overline{\mathcal{E}^{0.1,\gamma^{(5)}}_{\patch,1}}$}{ 
        \begin{overpic}[width=0.235\textwidth,tics=5]{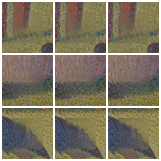} 
            \setlength{\tabcolsep}{0pt}
            \put (-2,-10){\small
            \begin{tabularx}{0.24\textwidth}{ 
                  >{\centering\arraybackslash}X
                  >{\centering\arraybackslash}X
                  >{\centering\arraybackslash}X
                  }
                 $\sigma=3$ & $\sigma=10$ & $\sigma=100$
            \end{tabularx}}
        \end{overpic}}
        \caption{Inpainting using models with Gaussian derivatives at scale $s=5$.}
        \label{fig:ex_2_result_b}
    \end{subfigure}
    \begin{subfigure}[t]{0.49\linewidth}
        \centering
        \setlength\abovecaptionskip{1.4\baselineskip}
        \begin{overpic}[width=\textwidth,tics=5]{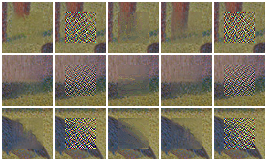} 
            \put (8,62){\small $\mathcal{E}^{1}_\patch$}
            \put (26,62){\small $\mathcal{E}^{\gamma^{(5)}}_1$}
            \put (43,62){\small $\mathcal{E}^{0.1,\gamma^{(5)}}_1$}
            \put (63,62){\small $\overline{\mathcal{E}^{0.1,\gamma^{(5)}}_{\patch,1}}$}
            \put (7,-5)   {\small $u_1$}
            \put (27,-5)  {\small $u_2$}
            \put (48,-5)  {$u_3$}
            \put (66,-5)  {$u_4$}
            \put (80,-5)  {\small$.1u_1{\scriptsize+}.9u_2$}
        \end{overpic} 
    	\caption{Comparison of five related models using $15\times 15$ kernels $\gamma(\bfh)$ with $\sigma=10$.}
    	\label{fig:ex_2_result_c}
	\end{subfigure}
	\:
	\begin{subfigure}[t]{0.49\linewidth}
        \centering
        \setlength\abovecaptionskip{1.5\baselineskip}
        \begin{overpic}[width=\textwidth,tics=5]{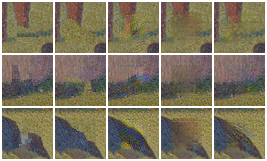} 
            \put (4,-5)   {\small Cr    \cite{Criminisi2004}}
            \put (22,-5)  {\small Ne    \cite{Newson2017}}
            \put (41,-5)  {\small cntx  \cite{Yu_2018_CVPR}}
            \put (63,-5)  {\small ec    \cite{nazeri2019}}
            \put (80,-5)  {\small gmcnn \cite{wang2018image}}
        \end{overpic}
    	\caption{Inpainting with the state-of-the-art exemplar-based and deep generative methods.}
    	\label{fig:ex_2_result_d}
	\end{subfigure}
    \caption[Inpainting of the image in Example 2.]{Inpainting of the image in Example 2.}
    \label{fig:ex_2_result}
\end{figure}

As expected, the nonlocal means algorithm is the best at interpolating image texture and less successful at interpolating non-trivial geometric structures.
At the same time, the nonlocal Poisson and $\gamma$-Poisson methods lead to the overly smoothed solutions and completely fail at reconstructing edges-like structures.
However, such behavior is common for highly textured images as their derivatives are distributed nearly uniformly in the vicinity of each pixel and hence their ``patch averaged" values do not contain useful information. 
This is clearly seen from the image of $\nabla_x u$ in Figure \ref{fig:ex_2_setup}; the same figure shows the Gaussian derivative $\nabla_x u^{(5)}$ defined~as
$$
\nabla_x^{(5)} u = \nabla_x \left(G^{(5)} * u\right),
\qquad\text{where}\quad
G^{(s)}(\bfx) = \frac{1}{2\pi s} \exp\left( -\frac{\|\bfx\|^2}{2s^2} \right)
$$
and $G^{(s)} * u$ is a smoothed image at scale $s$.
One can see that the high-frequency textures have been filtered out making the edge-like structures more pronounced.
Using the associative property of convolutions, we define the corresponding kernels $g_{\square^{(s)}}$ of the Gaussian derivatives at scale $s$ as
$$
g_{\square^{(s)}} * u = g_{\square} * \left(G^{(s)} * u\right) = \left(g_{\square} * G^{(s)}\right) * u.
$$
In our calculations, we used the finite Gaussian kernels truncated at $0.67$ standard deviations which corresponds to one half of their probability measure.

Figure \ref{fig:ex_2_result_b} illustrates the inpainting results for the models $\mathcal{E}^{0.01,\nabla^{(5)}}_\patch$, $\overline{\mathcal{E}^{0.01,\nabla^{(5)}}_{\patch,\patch}}$, $\mathcal{E}^{0.1,\gamma^{(5)}}_1$ and $\overline{\mathcal{E}^{0.1,\gamma^{(5)}}_{\patch,1}}$ using Gaussian versions of the local and nonlocal gradient operators at scale $s=5$.
Note that it was necessary to use $\mathcal{E}^{\alpha,\square^{(5)}}$ and $\overline{\mathcal{E}^{\alpha,\square^{(5)}}}$ with $\alpha\neq 0$ in this case to avoid the ill-posedness of the problem due to lossy Gaussian smoothing (compare $\mathcal{E}^{\gamma^{(5)}}_1$ and $\mathcal{E}^{0.1,\gamma^{(5)}}_{1,1}$ in Figure~\ref{fig:ex_2_result_c}).
It is clear that all four models in Figure \ref{fig:ex_2_result_b} resolve the edge-like structures much better than their related models in Figure \ref{fig:ex_2_result_a}.
The overall resolution of geometric structures is most accurate for the solutions based on the models with nonlocal gradients.
The textures are also resolved pretty well by all models except $\mathcal{E}^{0.1,\gamma^{(5)}}_1$.
The reason is trivial since this model is using $15\times 15$ kernels $\gamma(\bfh)$ of the nonlocal gradient operator and patches of size just $1\times 1$.
This resolves the color information correctly, but is not enough for the texture reconstruction.
At the same time, the mixture model $\overline{\mathcal{E}^{0.1,\gamma^{(5)}}_{\patch,1}}$ can easily combine $15\times 15$ kernels $\gamma(\bfh)$ in $\mathcal{E}^{\gamma^{(5)}}_1$ with $15\times 15$ patches $\patch(\bfh)$ in $\mathcal{E}^1_\patch$.
Finally, it should be noticed that the solutions corresponding to the less localized kernels in $\mathcal{E}^{0.1,\gamma^{(5)}}_1$, $\overline{\mathcal{E}^{0.1,\gamma^{(5)}}_{\patch,1}}$ are less smooth and hence more accurate as was expected.

Figure \ref{fig:ex_2_result_c} collects comparisons of five related models which share numerical values for some of their parameters.
Three results are worth mentioning: 1) the model $\mathcal{E}^{\gamma^{(5)}}_1$ with the solution $u_2$ is unstable justifying the use of additional regularization in $\mathcal{E}^{0.1,\gamma^{(5)}}_1$, 2) the mixture model $\overline{\mathcal{E}^{0.1,\gamma^{(5)}}_{\patch,1}}$ has better texture resolution than $\mathcal{E}^{0.1,\gamma^{(5)}}_1$, and 3) the solution $u_4$ of the mixture model is not a weighted average of the corresponding solutions $u_1$ and $u_2$.

Finally, Figure \ref{fig:ex_2_result_d} illustrates inpainting with the state-of-the-art exemplar and deep generative models trained on Places2 dataset.
It is seen that the Generative Multi-column Convolutional Neural Network (gmcnn) model shows the best result which nevertheless does not outperform the best results of the proposed model.

\paragraph{Example 3}

\begin{figure}[t]
    \centering
    \stackinset{c}{0pt}{b}{-15pt}{\small Original image}{\includegraphics[width=0.24\textwidth]{example_3/image.png}}
    \stackinset{c}{0pt}{b}{-15pt}{\small Available exemplars}{\includegraphics[width=0.24\textwidth]{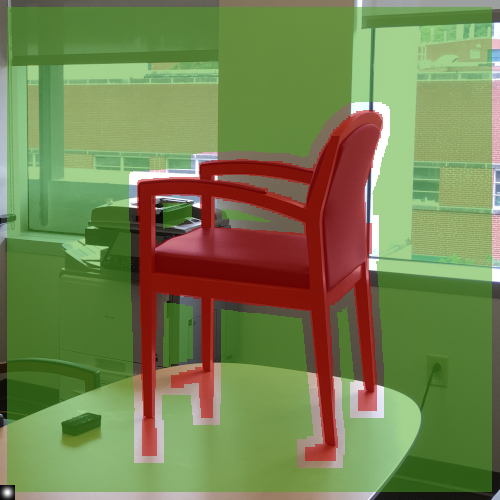}}
    \stackinset{c}{0pt}{b}{-15pt}{\small Edge completion $E(\bfx)$}{\includegraphics[width=0.24\textwidth]{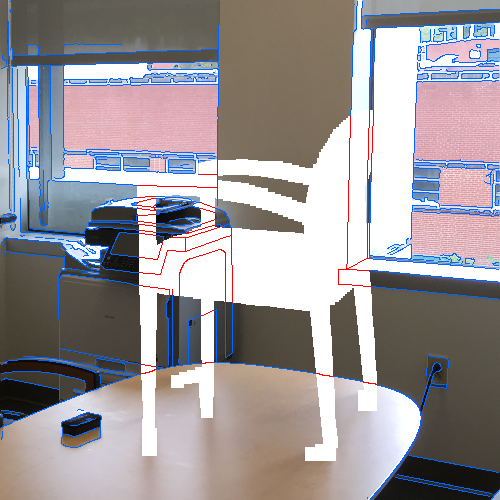}}
    \stackinset{c}{0pt}{b}{-15pt}{\small Coefficient $\lambda(\bfx)$}{\includegraphics[width=0.24\textwidth]{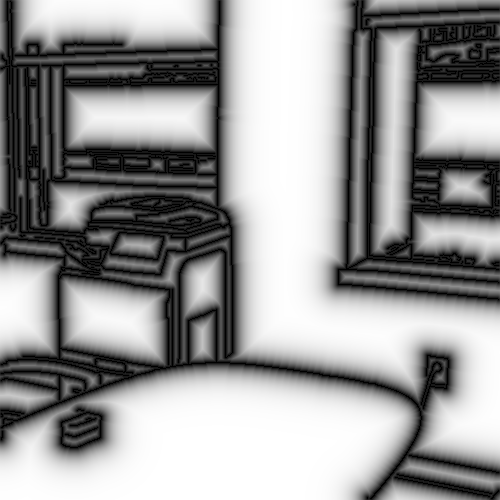}}
    \caption{Setup of Example 3.}
    \label{fig:ex_3_setup}
\end{figure} 

\begin{figure}[t]
    \centering
    \begin{subfigure}[t]{\linewidth}
    	\centering
    \stackinset{c}{0pt}{t}{-15pt}{\small \Centerstack{Local Poisson}}{\includegraphics[width=0.24\textwidth]{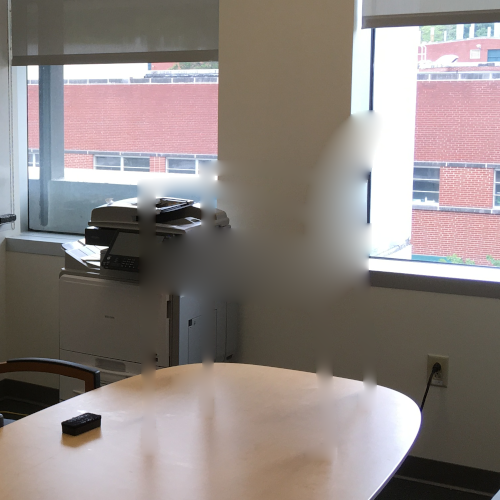}}
    	\stackinset{c}{0pt}{t}{-15pt}{\small \Centerstack{Nonlocal Poisson}}{\includegraphics[width=0.24\textwidth]{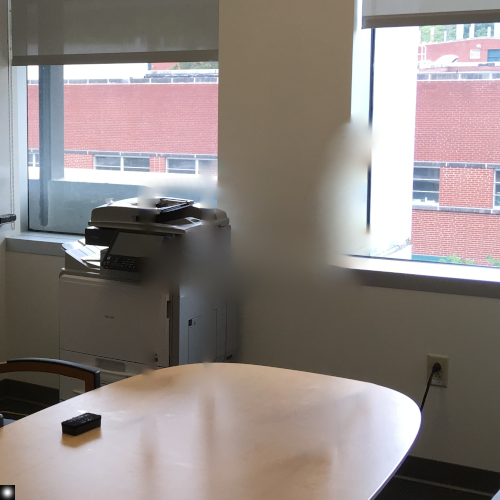}}
    	\stackinset{c}{0pt}{t}{-30pt}{\small \Centerstack{Local Poisson \\ with edge completion}}{\includegraphics[width=0.24\textwidth]{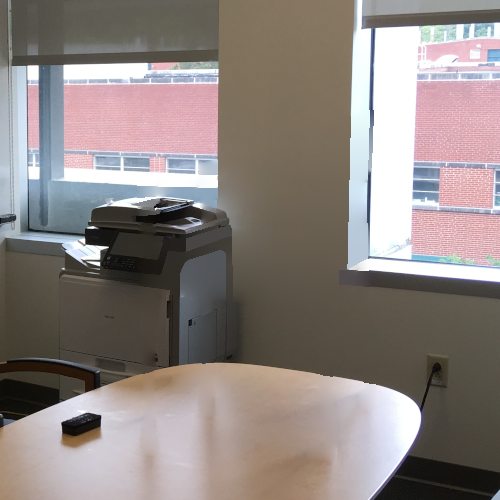}}
    	\stackinset{c}{0pt}{t}{-30pt}{\small \Centerstack{Nonlocal Poisson \\ with edge completion}}{\includegraphics[width=0.24\textwidth]{example_3/nonloc_harmonic_edges.png}}
        \caption{Local vs nonlocal Poisson inpainting without and with manual edge completion in Figure~\ref{fig:ex_3_setup}.}
        \label{fig:ex_3_result_a}
    \end{subfigure}
    \begin{subfigure}[t]{.48\linewidth}
    	\centering
        \includegraphics[width=0.49\textwidth]{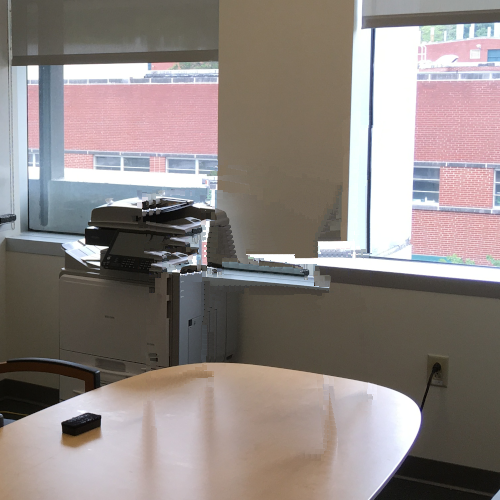}
        \includegraphics[width=0.49\textwidth]{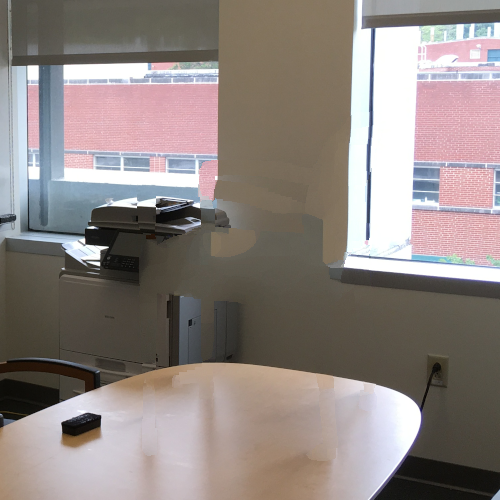}
        \caption{Exemplar-based inpainting by Criminisi et al. \cite{Criminisi2004} and Newson et al. \cite{Newson2017}.}
    	\label{fig:ex_3_edges_result_b}
    \end{subfigure}
    \quad
    \begin{subfigure}[t]{.48\linewidth}
    	\centering
        \includegraphics[width=0.49\textwidth]{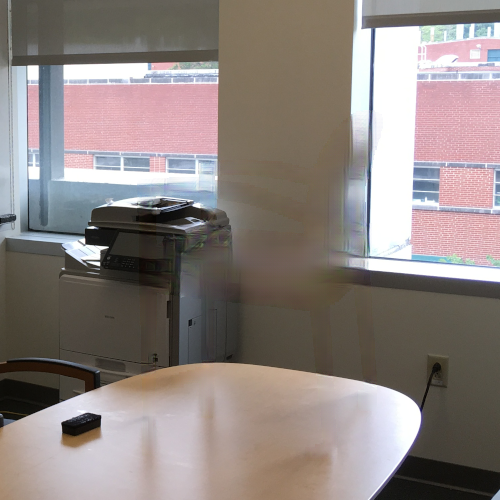}
        \includegraphics[width=0.49\textwidth]{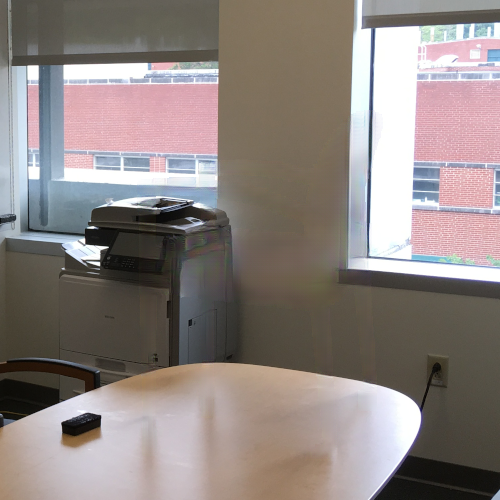}
        \caption{EdgeConnect \cite{nazeri2019} inpainting without and with manual edge completion in Figure~\ref{fig:ex_3_setup}.}
        \label{fig:ex_3_edges_result_c}
    \end{subfigure}
    \begin{subfigure}[t]{\linewidth}
    	\centering
    \stackinset{c}{0pt}{b}{-15pt}{\small Iizuka  et al. \cite{IizukaSIGGRAPH2017}}{\includegraphics[width=0.24\textwidth]{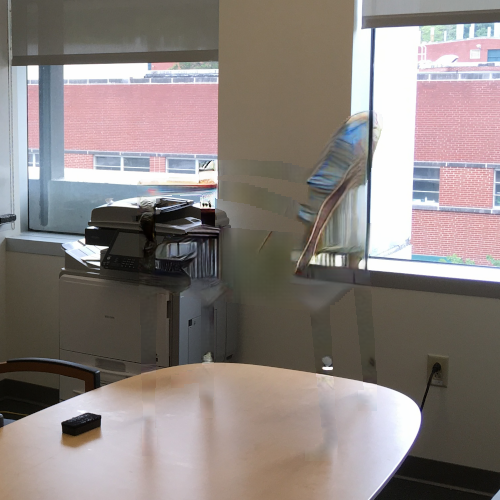}}
    	\stackinset{c}{0pt}{b}{-15pt}{\small Liu et al.   \cite{liu2018partialinpainting}}{\includegraphics[width=0.24\textwidth]{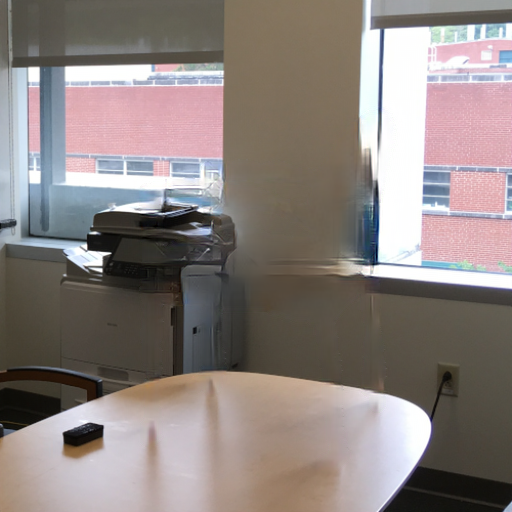}}
    	\stackinset{c}{0pt}{b}{-15pt}{\small Wang et al. \cite{wang2018image}}{\includegraphics[width=0.24\textwidth]{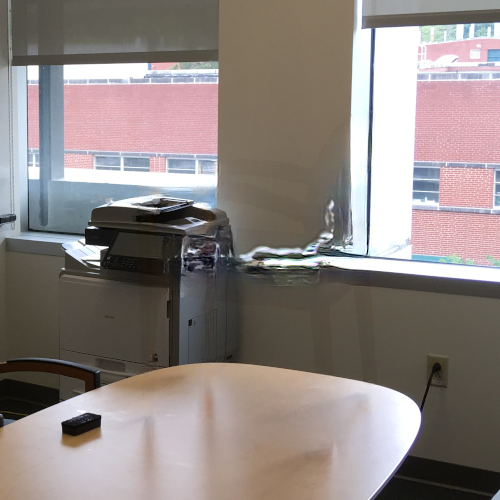}}
    	\stackinset{c}{0pt}{b}{-15pt}{\small Yu et al. \cite{Yu_2018_CVPR}}{\includegraphics[width=0.24\textwidth]{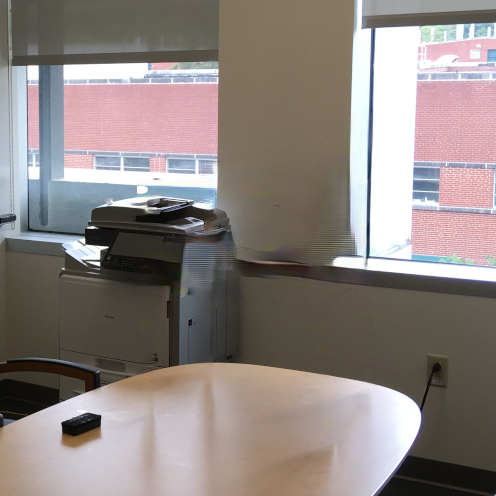}}
        \caption{State-of-the-art deep generative inpainting models trained on Places2 dataset.}
        \label{fig:ex_3_edges_result_d}
    \end{subfigure}
    \caption{Inpainting of the image in Example 3.}
    \label{fig:ex_3_result}
\end{figure} 

The aim of this example is to demonstrate capabilities of the algorithm in reconstructing geometric structures.
Figure \ref{fig:ex_3_setup} shows the setup of Example 3; one can see that the image is almost cartoon-like with only a few textured regions.
Following terminology of the previous examples, convolutional kernels $g_{\nabla}$, $g_{\nabla_y}$ representing gradient of the image seem to be a natural choice in this case.
When $\lambda_{\nabla_x}=\lambda_{\nabla_y}=1$, the image update step of the algorithm is thus given by the Dirichlet problem for the Poisson equation with a nonlocal forcing term.
The first two figures in Figure \ref{fig:ex_3_result_a} illustrate the obtained results for the standard local and nonlocal Poisson methods. 
The local solution was also used to initialize the corresponding nonlocal method.
It is clear that the nonlocal approach leads to better propagation of edges across small holes (e.g., missing edges of a table or the wall over the printer) but still performs far from perfect in larger regions (e.g., the windows) and regions with structures not present elsewhere in the image (e.g., the printer).

In order to improve the result, we can supplement the algorithm with a hint in the form of the manual edge completion $E(\bfx)$ in Figure \ref{fig:ex_3_setup}.
This permits one to consider a nonlocal patch distance metric \eqref{eq:our_dist} with anisotropic coefficient in \eqref{eq:aniso_lambda}.
The Euler-Lagrange equation \eqref{eqn:our_bdyproblem} in this case describes an anisotropic diffusion with $\lambda(\bfx)$ controlling the intensity of diffusion within the image.
The last two figures in Figure~\ref{fig:ex_3_result_a} show the results of the local and nonlocal inpaintings using this approach.
The local method again has been used as an initialization for the nonlocal model.
Both methods are successful but it is noticeable that the nonlocal technique produces crisper edges and more natural colors.
The distinction is much more apparent for textured images like the one in Figure \ref{fig:ex_1_edges}.

For the sake of comparison, we have also considered inpainting with the state-of-the-art exemplar (Figure~\ref{fig:ex_3_edges_result_b}) and deep generative models trained on the Places2 dataset (Figure~\ref{fig:ex_3_edges_result_d}).
The EdgeConnect generative model in \cite{nazeri2019}, which allows user-supplied edge completion, is shown separately in Figure \ref{fig:ex_3_edges_result_c}.
The results in Figure \ref{fig:ex_3_result} show the superiority of the proposed nonlocal method to other presented methods.

\paragraph{Example 4}\label{example:4}

\begin{figure}
    \centering
    \stackinset{c}{0pt}{b}{-15pt}{\small Original image}{\includegraphics[width=0.24\textwidth]{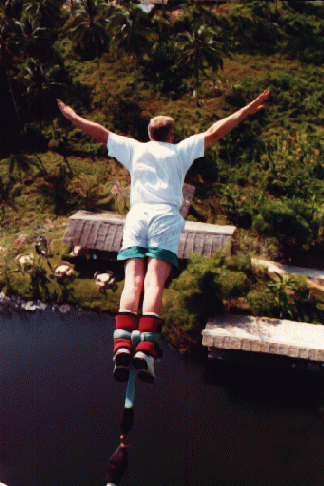}}
    \stackinset{c}{0pt}{b}{-15pt}{\small Available exemplars}{\includegraphics[width=0.24\textwidth]{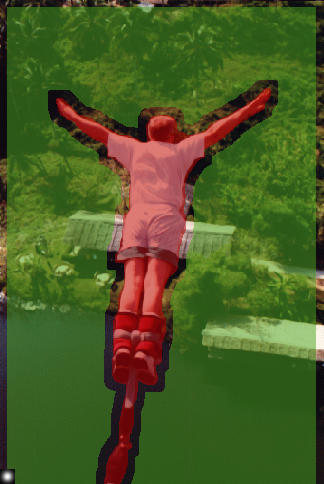}}
    \stackinset{c}{0pt}{b}{-15pt}{\small Manual edge completion}{\includegraphics[width=0.24\textwidth]{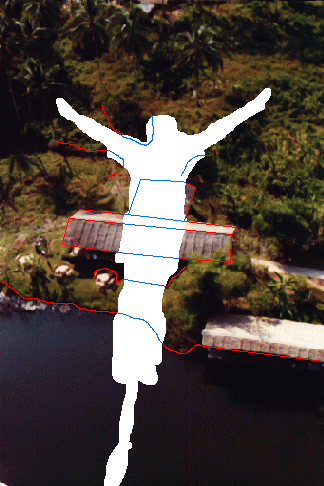}}
    \stackinset{c}{0pt}{b}{-15pt}{\small Model masks}{
    \begin{overpic}[width=0.24\textwidth,tics=5]{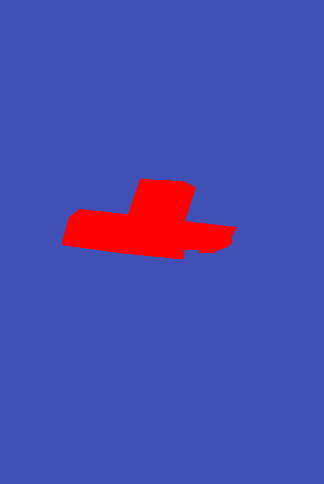} 
        \put (25,85) {\small\color{white} model $1$}
        \put (19,51) {\small\color{white} \rotatebox{-7}{model $2$}}
    \end{overpic}}
    \caption{Setup of Example 4.}
    \label{fig:ex_4_setup}
\end{figure} 

\begin{figure}
    \begin{subfigure}{\linewidth}
        \centering
        \stackinset{c}{0pt}{t}{-15pt}{\small Local Poisson}{\includegraphics[width=0.24\textwidth]{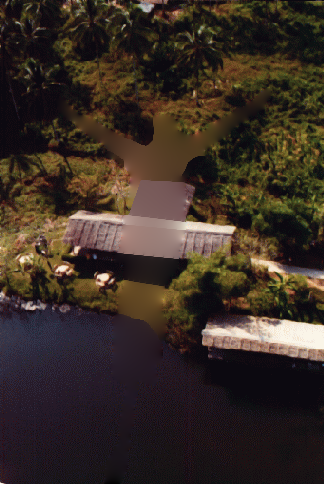}}
        \stackinset{c}{0pt}{t}{-15pt}{\small Nonlocal Poisson}{\includegraphics[width=0.24\textwidth]{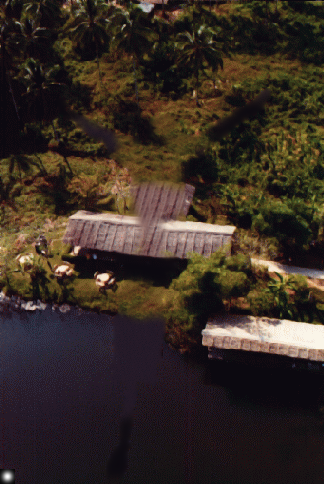}}
        \stackinset{c}{0pt}{t}{-30pt}{\footnotesize \Centerstack{Model 1: nonlocal means \\ Model 2: nonlocal means}}{\includegraphics[width=0.24\textwidth]{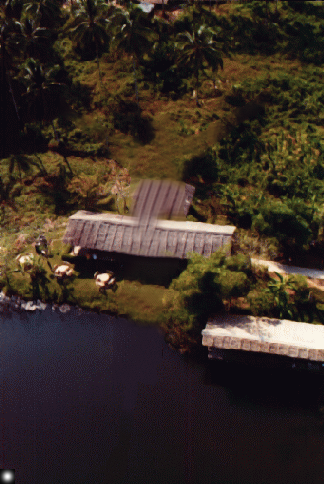}}
        \stackinset{c}{0pt}{t}{-30pt}{\footnotesize \Centerstack[l]{Model 1: nonlocal means \\ Model 2: nonlocal Poisson}}{\includegraphics[width=0.24\textwidth]{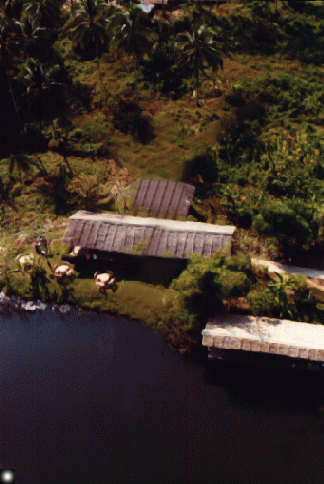}}
        \caption{Inpainting without and with manual edge completion and model masks in Figure~\ref{fig:ex_4_setup}.}
        \label{fig:ex_4_result_a}
    \end{subfigure}
    \begin{subfigure}{\linewidth}
        \centering
        \stackinset{c}{0pt}{b}{-15pt}{\small Criminisi et al. \cite{Criminisi2004}}{\includegraphics[width=0.24\textwidth]{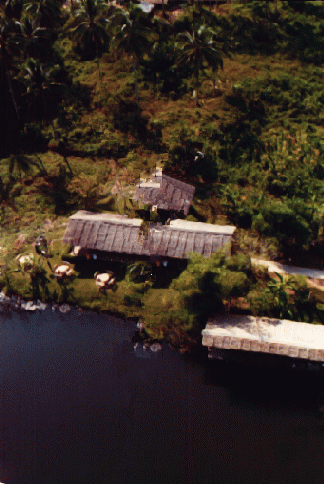}}
        \stackinset{c}{0pt}{b}{-15pt}{\small Newson et al. \cite{Newson2017}}{\includegraphics[width=0.24\textwidth]{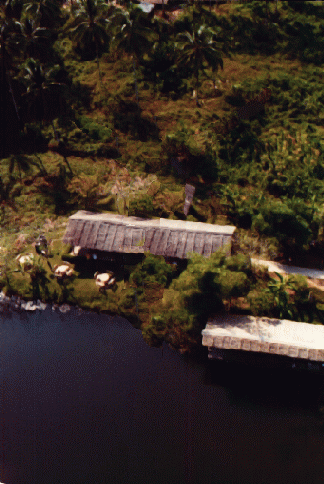}}
        \stackinset{c}{0pt}{b}{-30pt}{\small \Centerstack{Fedorov et al. \cite{Fedorov_ipol2015}\\ nlmeans}}{\includegraphics[width=0.24\textwidth]{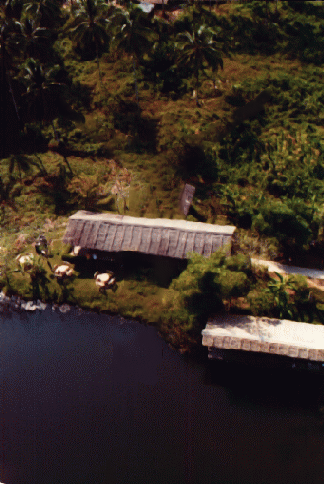}}
        \stackinset{c}{0pt}{b}{-30pt}{\small \Centerstack{Fedorov et al. \cite{Fedorov_ipol2015}\\ nlpoisson ($\lambda=0.2$)}}{\includegraphics[width=0.24\textwidth]{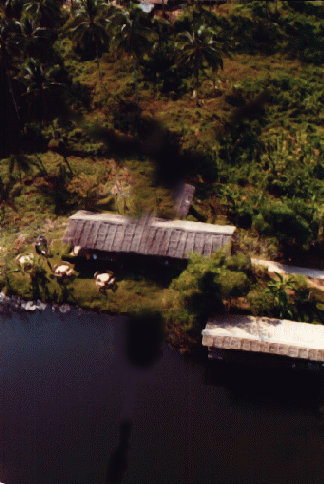}}
        \caption{State-of-the-art exemplar-based inpainting using $9\times 9$ patches and multiscale initialization.}
        \label{fig:ex_4_result_b}
    \end{subfigure}
    \begin{subfigure}[t]{\linewidth}
    	\centering
    	\stackinset{c}{0pt}{b}{-15pt}{\small Iizuka  et al. \cite{IizukaSIGGRAPH2017}}{\includegraphics[width=0.19\textwidth]{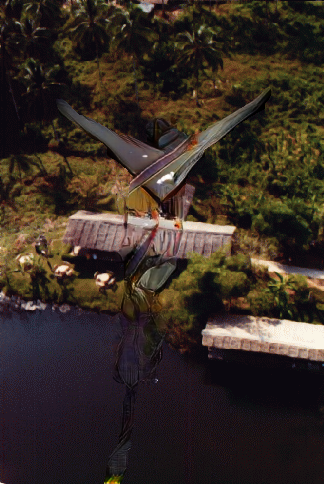}}
    	\stackinset{c}{0pt}{b}{-15pt}{\small Liu et al. \cite{liu2018partialinpainting}}{\includegraphics[width=0.19\textwidth]{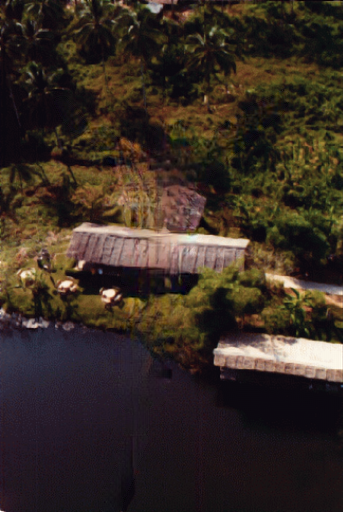}}
    	\stackinset{c}{0pt}{b}{-15pt}{\small Wang et al. \cite{wang2018image}}{\includegraphics[width=0.19\textwidth]{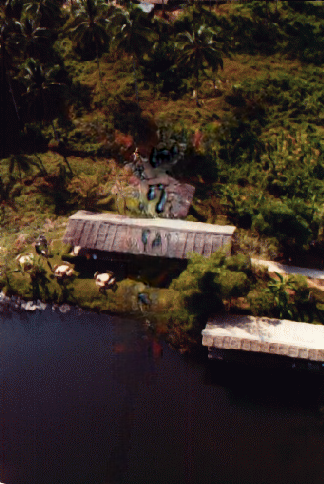}}
    	\stackinset{c}{0pt}{b}{-15pt}{\small Yu et al. \cite{Yu_2018_CVPR}}{\includegraphics[width=0.19\textwidth]{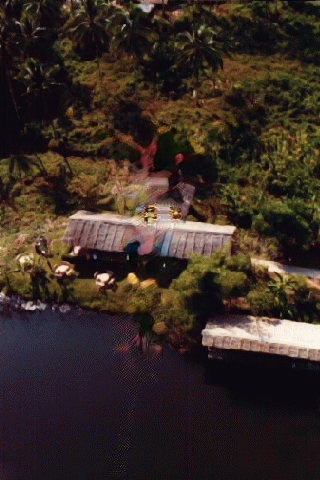}}
    	\stackinset{c}{0pt}{b}{-15pt}{\small EdgeConnect \cite{nazeri2019}}{\includegraphics[width=0.19\textwidth]{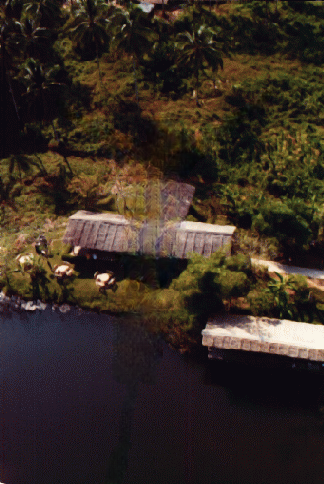}}
        \caption{State-of-the-art deep generative inpainting models trained on Places2 dataset.}
        \label{fig:ex_4_result_c}
    \end{subfigure}
    \caption{Inpainting of the image in Example 4.}
    \label{fig:ex_4_result}
\end{figure}

Our final example demonstrates the flexibility of the proposed algorithm in controlling its behavior.
For this purpose, consider the image in Figure \ref{fig:ex_4_setup} and its inpainting in Figure \ref{fig:ex_4_result_b} obtained with a multiscale initialization strategy using a $15\times 15$ Gaussian patch with $\sigma=5$.
It is seen that the state-of-the-art exemplar methods demonstrate plausible results in the terrain area but are only somewhat successful at reconstructing the missing roof of the house.

As a first improvement attempt, we enforce the edge completion and calculate the anisotropic coefficient $\lambda_{\nabla}(\bfx)$ in \eqref{eq:aniso_lambda} with $\lambda_a=0.1$ and $\tau=10$.
The first two figures in Figure \ref{fig:ex_4_result_a} show the inpainting results of the the local and nonlocal Poisson models illustrating the relative success of the method.
As before, the local solution was used to initialize the nonlocal methods.
While the desired geometry is resolved much better in comparison to conventional methods in Figure \ref{fig:ex_4_result_b}, interpolation of the textured regions (grass and water) is not quite acceptable.

To improve the result even further, we separate the inpainting regions as shown in the last figure of Figure \ref{fig:ex_4_setup} and fix the nonlocal means method as model 1 in the terrain region.
The last two figures of Figure \ref{fig:ex_4_result_b} illustrates the inpainting results for two choices of model 2 (house): the nonlocal means and nonlocal Poisson.
As one can see, this leads to crisper edges and better resolution of textures as desired.
In our opinion, the proposed technique provides an appealing solution to the inpainting problem, especially when compared to solutions of the state-of-the-art exemplar and deep generative models in Figures \ref{fig:ex_4_result_b}-\ref{fig:ex_4_result_c}.

\section{Conclusion and future work}\label{sec:conclusion}

In this work we presented an image inpainting model which is free to depend on a variety of features. 
The model is capable of propagating structure within the inpainting region as well as exploiting self-similarity in intact portions of the image to recover texture. 
The obtained numerical results support our findings and illustrate superiority of the proposed technique in comparison to similar methods.

We see several opportunities for potential improvements of the presented model.
For example, we derived the anisotropic coefficient in the patch distance metric \eqref{eq:our_dist} using manual edge completion and handcrafted convolutions.
It would be interesting to explore automated techniques for edge reconstruction and data-driven selection of convolutional operators in the spirit of deep learning approaches.
Additionally, the metric \eqref{eq:our_dist} is quadratic. Exploring alternative norms could lead to superior results. For example, it is known that TV like formulations are more natural for images and frequently lead to sharper transition regions. Finally, the inclusion of convolutions into the metric \eqref{eq:our_dist} results in a boundary value problem for the unknown parts of the image.
The well-posedness of such problems should be also analyzed.
We would also like to consider more general (potentially nonlinear) operators in addition to convolutions.
We intend to study these questions in our future works.

\section*{Acknowledgements}

This material is based upon work supported in part by: the U.S.
Department of Energy, Office of Science, Early Career Research Program under award number 
ERKJ314; U.S. Department of Energy, Office of Advanced Scientific Computing Research under 
award numbers ERKJ331 and ERKJ345; the National Science Foundation, Division of Mathematical 
Sciences, Computational Mathematics program under contract number DMS1620280; and the 
Behavioral Reinforcement Learning Lab at Lirio LLC.

We would also like to thank the anonymous reviewers for their valuable comments which let to the improvement of the manuscript.

\appendix
\section{Derivation of the image updating equation of Algorithm~\ref{alg:our_algorithm}}\label{appendix:algorithm} 
In this section we derive the Euler-Lagrange equation \eqref{eqn:our_bdyproblem} of \eqref{eq:our_model}. We begin with the expanded form of the functional \eqref{eq:our_model}:
\begin{align*}
    \sum_{j=1}^{N_{\beta}} \sum_{i=1}^{N_{g}} \int\displaylimits_{\mathbb{R}^2} \int\displaylimits_{\mathbb{R}^2} \int\displaylimits_{\patch^j}
    &\indicator{\tilde{\inpdom}_*}(\bfz-\bfh) \indicator{\tilde{\inpdom}_*^c}(\hat{\bfz}-\bfh) w^j(\bfz-\bfh,\hat{\bfz}-\bfh)
    \\
    &\beta^j(\bfz-\bfh)\lambda_i^j(\bfz)\Big( (g_i*\bfu)(\bfz) - (g_i*\bfu)(\hat{\bfz}) \Big)^2 d\patch^j(\bfh) d\hat{\bfz} d\bfz,
\end{align*}
where $\indicator{A}(\bfx)$ is the indicator function of the set $A$ and we used the change of variables $\bfz=\bfx+\bfh$ and $\hat{\bfz}=\bfy+\bfh$. 
Since $w_j(\bfz-\bfh, \hat{\bfz}-\bfh) = 0$ for $\hat{\bfz} - \bfh \in \tilde{\mathcal{O}}_*$ by definition ({\em cf}. \eqref{eq:feature_weights}), 
we may remove $\indicator{\tilde{\inpdom}_*^c}(\hat{\bfz}-\bfh)$ from the integrand. 
In addition, since $\hat{\bfz} - \bfh \in \tilde{\mathcal{O}}_*^c$ implies $\hat{\bfz} \in \mathcal{O}_*^c$ by \eqref{def:Otiledstar}, we may reduce the limits of integration in $\hat{\bfz}$ to $\mathcal{O}_*^c$: 
\begin{align}\label{eqn:functionaltemp1}
    \mathcal{E}[\bfu] := \sum_{i,j} \int\displaylimits_{\mathbb{R}^2} \int\displaylimits_{\mathcal{O}_*^c} \int\displaylimits_{\patch_j}
    &\indicator{\tilde{\inpdom}_*}(\bfz-\bfh) w_j(\bfz-\bfh,\hat{\bfz}-\bfh)
    \\\nonumber
    &\beta^j(\bfz-\bfh)\lambda_i^j(\bfz)\Big( (g_i*\bfu)(\bfz) - (g_i*\bfu)(\hat{\bfz}) \Big)^2 d\patch^j(\bfh) d\hat{\bfz} d\bfz.
\end{align}
Next we notice the integral
\begin{align*} 
    \sum_{i,j} \int\displaylimits_{\inpdom_*^c} \int\displaylimits_{\inpdom_*^c} \int\displaylimits_{\patch^j}
    &\indicator{\tilde{\inpdom}_*}(\bfz-\bfh)  w^j(\bfz-\bfh,\hat{\bfz}-\bfh)
    \beta^j(\bfz-\bfh) \lambda_i^j(\bfz)\Big( (g_i*\bfu)(\bfz) - (g_i*\bfu)(\hat{\bfz}) \Big)^2 d\patch^j(\bfh) d\hat{\bfz} d\bfz,
\end{align*}
is constant in the image update step\footnote{Recall the weights $w$ are held constant in the image update step.}. 
To see this we simply note that by \eqref{def:Ostart}, we have for $\hat{\bfz}, \bfz \in \mathcal{O}_*^c$ that $g_i*(\bfu(\hat{\bfz})-\bfu(\bfz))$ is informed entirely by the region $\mathcal{O}^c$ and therefore unchanged in the image update step. Consequently, a minimizer of \eqref{eqn:functionaltemp1} is also a minimizer of the functional
\begin{align}\label{eqn:imageupdatefunctional1}
    \mathcal{E}[\bfu] &:= \sum_{i,j} \int\displaylimits_{\inpdom_*} \int\displaylimits_{\inpdom_*^c} \int\displaylimits_{\patch^j} \beta^j(\bfz-\bfh) w^j(\bfz-\bfh,\hat{\bfz}-\bfh)
    \lambda_i^j(\bfz)\Big( (g_i*\bfu)(\bfz) - (g_i*\bfu)(\hat{\bfz}) \Big)^2 d\patch^j(\bfh) d\hat{\bfz} d\bfz 
\end{align}
since \eqref{eqn:imageupdatefunctional1} and \eqref{eqn:functionaltemp1} only differ by a constant. Recalling the definitions \eqref{def:shorthandforimageupdate1} -- \eqref{def:shorthandforimageupdate3} and using the weight function in \eqref{eq:delta_weights}, we get
\begin{align*}
    &k^j(\bfz) = m^j(\bfz,\hat{\bfz}) = \int\limits_{\patch^j} \beta^j(\bfz-\bfh) d\patch^j(\bfh),
    \\
    &f_i^j(\bfz) = \int\displaylimits_{\patch^j} \beta^j(\bfz-\bfh) (g_i * \bfu) \big(\varphi^j(\bfz-\bfh)+\bfh\big) d\patch^j(\bfh)
\end{align*}
since $m^j(\bfz,\hat{\bfz})=0$ otherwise.

For $\hat{\bfz} \in \mathcal{O}_*^c$, we know $(g_i * \bfu)(\hat{\bfz})$ is entirely dependent on $\mathcal{O}^c$. Also recall the weighting functions $w^j$ are treated as constant in the image update step. 
Consequently, $m^j(\bfz,\hat{\bfz})$, $k^j(\bfz)$, and $f_i^j(\bfz)$ are constant in the image update step. Hence, by expanding the quadratic term and regrouping appropriately, the functional \eqref{eqn:imageupdatefunctional1} can be simplified as
\begin{align*}
    \mathcal{E}[\bfu] &:= 
    \sum_{i,j} \int\displaylimits_{\inpdom_*} \int\displaylimits_{\inpdom_*^c} m^j(\bfz,\hat{\bfz}) \lambda_i^j(\bfz) \Big( (g_i*\bfu)(\bfz) - (g_i*\bfu)(\hat{\bfz}) \Big)^2 d\hat{\bfz} d\bfz
    \\
    &= \sum_{i,j} \int\displaylimits_{\inpdom_*} k^j(\bfz) \lambda_i^j(\bfz) \left( (g_i*\bfu)(\bfz) - \frac{f_i^j(\bfz)}{k^j(\bfz)} \right)^2 d\bfz + const
\end{align*}

Hence the minimizer of \eqref{eqn:imageupdatefunctional1} is also a minimizer of the functional
\begin{align}\label{eqn:functionalbeforevariation}
    &\mathcal{E}[\bfu] :=
    \sum_{i,j} \int_{\inpdom_*} k^j(\bfz) \lambda_i^j(\bfz) \left( (g_i*\bfu)(\bfz) - \frac{f_i^j(\bfz)}{k^j(\bfz)} \right)^2 d\bfz.
\end{align}

Now that we have determined minimizers of \eqref{eq:our_model} in the image update step are minimizers of \eqref{eqn:functionalbeforevariation}, we will simply deal with \eqref{eqn:functionalbeforevariation} for the remainder of the derivation. In order to minimize \eqref{eqn:functionalbeforevariation}, we consider its first variation. Since $\bfu$ is constant in $\mathcal{O}^c$, in the calculation of the first variation of \eqref{eqn:functionalbeforevariation} we consider a function $\bfv(\bfz)$ which is identically zero in $\mathcal{O}^c$. We find 
\begin{align}\label{eqn:firstvariation}
    \frac{1}{2}\left. \frac{d}{d \epsilon} \mathcal{E}[\bfu+\epsilon \bfv] \right|_{\epsilon=0}
    &= \sum_{i,j} \int_{\mathbb{R}^2} \indicator{\inpdom_*}(\bfz) \lambda_i^j(\bfz) \big( k^j (g_i*\bfu) - f_i^j \big)(\bfz) (g_i*\bfv)(\bfz) d\bfz 
    \\\nonumber
    &= \sum_{i,j} \int_{\mathbb{R}^2} \left[ \overline{g}_i * \Big( \indicator{\inpdom_*} \lambda_i^j \big( k^j (g_i*\bfu) - f_i^j \big) \Big) \right] (\bfz) \bfv(\bfz) d\bfz 
    \\\nonumber
    &= \sum_{i,j} \int\limits_{\mathcal{O}} \left[ \overline{g}_i * \Big( \indicator{\inpdom_*} \lambda_i^j \big( k^j (g_i*\bfu) - f_i^j \big) \Big) \right] (\bfz) \bfv(\bfz) d\bfz 
    \\\nonumber
    &= \sum_{i,j} \int\limits_{\inpdom} \Big[ \overline{g}_i*\big( k^j \lambda_i^j (g_i*\bfu) \big) - \overline{g}_i * (\lambda_i^j f_i^j) \Big](\bfz) \bfv(\bfz) d\bfz,
\end{align}
where $\overline{g}_i(\bft) = g_i(-\bft)$ is the kernel of the adjoint convolution operator. We used the fact that $v(z)=0$ for $z\in\inpdom^c$ for the third equality in \eqref{eqn:firstvariation} and for the fourth equality in \eqref{eqn:firstvariation} we used~\eqref{def:Ostart} and the fact that $\supp(\overline{g}_i) = \supp(g_i)$ since $g_i$ has symmetric support. Hence, the Euler-Lagrange equations are given by \eqref{eqn:our_bdyproblem}. 

\newpage
\bibliographystyle{siamplain}
\bibliography{biblio}

\end{document}